\documentclass[10pt,twocolumn,letterpaper]{article}

\pdfoutput=1

\usepackage{cvpr}
\usepackage{times}
\usepackage{graphicx}
\usepackage{amsmath}
\usepackage{amssymb}

\usepackage[ruled]{algorithm2e}

\usepackage{amsthm}

\usepackage[caption = false]{subfig}

\usepackage{multirow}

\usepackage[pagebackref=true,breaklinks=true,letterpaper=true,colorlinks,bookmarks=false]{hyperref}

\usepackage{cleveref}
\crefformat{section}{\S#2#1#3} 
\crefformat{subsection}{\S#2#1#3}
\crefformat{subsubsection}{\S#2#1#3}

\usepackage[T1]{fontenc}

\DeclareMathOperator*{\argmin}{arg\,min}

\newtheorem{corollary}{Corollary}

\theoremstyle{definition}
\newtheorem{remark}{Remark}

\newcommand{\lfc}{\textsf{LFC}}
\newcommand{\hfc}{\textsf{HFC}}

\newcommand{\mn}{\textsf{M\textsubscript{natural}}}
\newcommand{\ms}{\textsf{M\textsubscript{shuffle}}}
\newcommand{\ma}{\textsf{M\textsubscript{adversarial}}}

\newcommand\blfootnote[1]{%
  \begingroup
  \renewcommand\thefootnote{}\footnote{#1}%
  \addtocounter{footnote}{-1}%
  \endgroup
}

\cvprfinalcopy 

\def\cvprPaperID{10190} 

\ifcvprfinal\fi
\begin{document}

\title{High-frequency Component Helps Explain \\the Generalization of Convolutional Neural Networks}

\author{Haohan Wang, Xindi Wu, Zeyi Huang, Eric P. Xing\\
School of Computer Science\\
Carnegie Mellon University\\
{\tt\small \{haohanw,epxing\}@cs.cmu.edu, \{xindiw,zeyih\}@andrew.cmu.edu}
}

\maketitle

\begin{abstract}
   We investigate the relationship between the frequency spectrum of image data and the generalization behavior of convolutional neural networks (CNN). 
We first notice CNN's ability in capturing the high-frequency components of images. 
These high-frequency components are almost imperceptible to a human. 
Thus the observation leads to multiple hypotheses that are related to the generalization behaviors of CNN, 
including a potential explanation for adversarial examples, 
a discussion of CNN's trade-off between robustness and accuracy, 
and some evidence in understanding training heuristics. 
\end{abstract}

\thispagestyle{empty} 

\section{Introduction}
Deep learning has achieved many recent advances in predictive modeling in various tasks,
but the community has nonetheless become alarmed by the unintuitive generalization behaviors of neural networks, such as the capacity in memorizing label shuffled data \cite{zhang2016understanding} and the vulnerability towards adversarial examples \cite{szegedy2013intriguing,goodfellow2015explaining}

To explain the generalization behaviors of neural networks, many theoretical breakthroughs have been made progressively, including studying the properties of stochastic gradient descent \cite{keskar2016largebatch}, different complexity measures \cite{neyshabur2017exploring}, generalization gaps \cite{schmidt2018adversarially}, and many more from different model or algorithm perspectives \cite{kawaguchi2017generalization,mahloujifar2018curse,bubeck2018adversarial,shamir2019simple}. 

\begin{figure}[]
    \centering
    \includegraphics[width=0.45\textwidth]{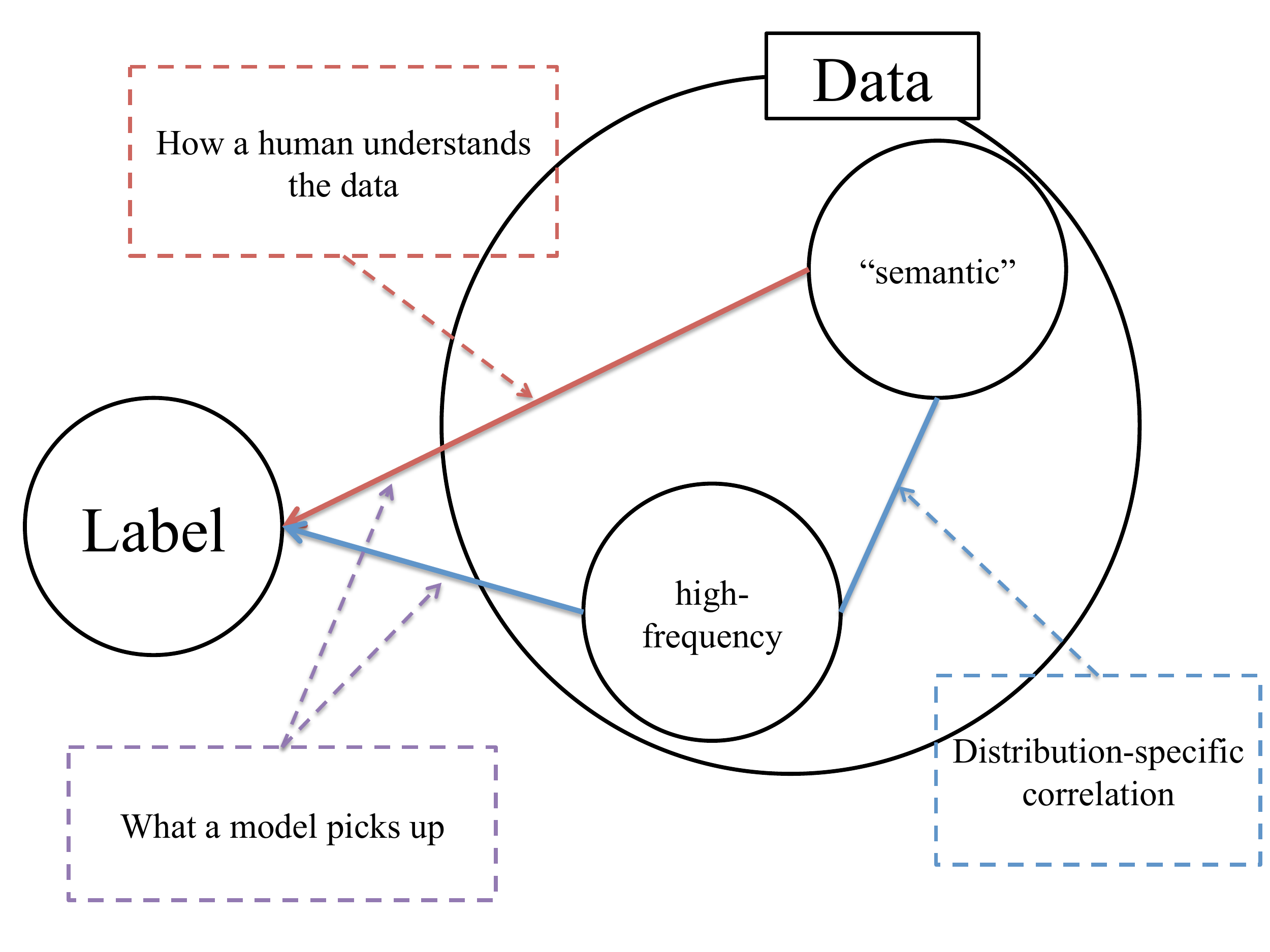}
    \caption{The central hypothesis of our paper: within a data collection, there are correlations between the high-frequency components and the ``semantic'' component of the images. As a result, the model will perceive both high-frequency components as well as the ``semantic'' ones, leading to generalization behaviors counter-intuitive to human (\textit{e.g.}, adversarial examples).}
    \label{fig:main}
\end{figure}

In this paper, inspired by previous understandings that convolutional neural networks (CNN) can learn from confounding signals \cite{wang2017select} and superficial signals \cite{jo2017measuring,geirhos2018imagenettrained,wang2018learning},
we investigate the generalization behaviors of CNN from a data perspective. 
Together with \cite{ilyas2019adversarial}, we suggest that the unintuitive generalization behaviors of CNN as a direct outcome of the perceptional disparity between human and models (as argued by Figure~\ref{fig:main}):
\emph{CNN can view the data at a much higher granularity than the human can}.

However, different from \cite{ilyas2019adversarial}, 
we provide an interpretation of this high granularity of the model's perception:
\emph{CNN can exploit the high-frequency image components that are not perceivable to human}. 

\begin{figure*}
\centering 
\subfloat[A sample of frog]{
\begin{tabular}[b]{ccc}%
  \includegraphics[width=.07\linewidth]{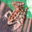}
  \includegraphics[width=.07\linewidth]{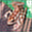}
  \includegraphics[width=.07\linewidth]{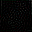} \\
  \includegraphics[width=.07\linewidth]{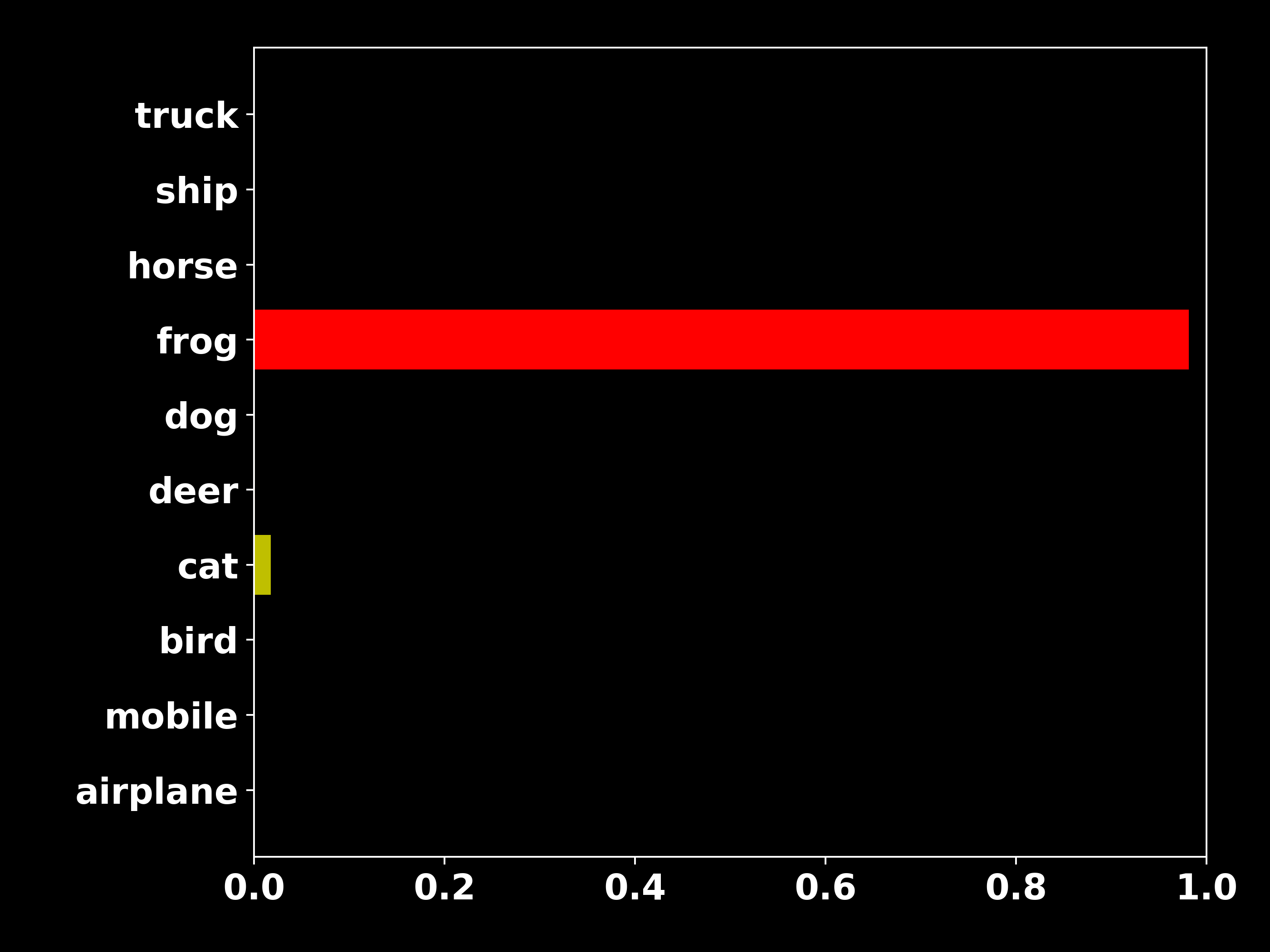}
  \includegraphics[width=.07\linewidth]{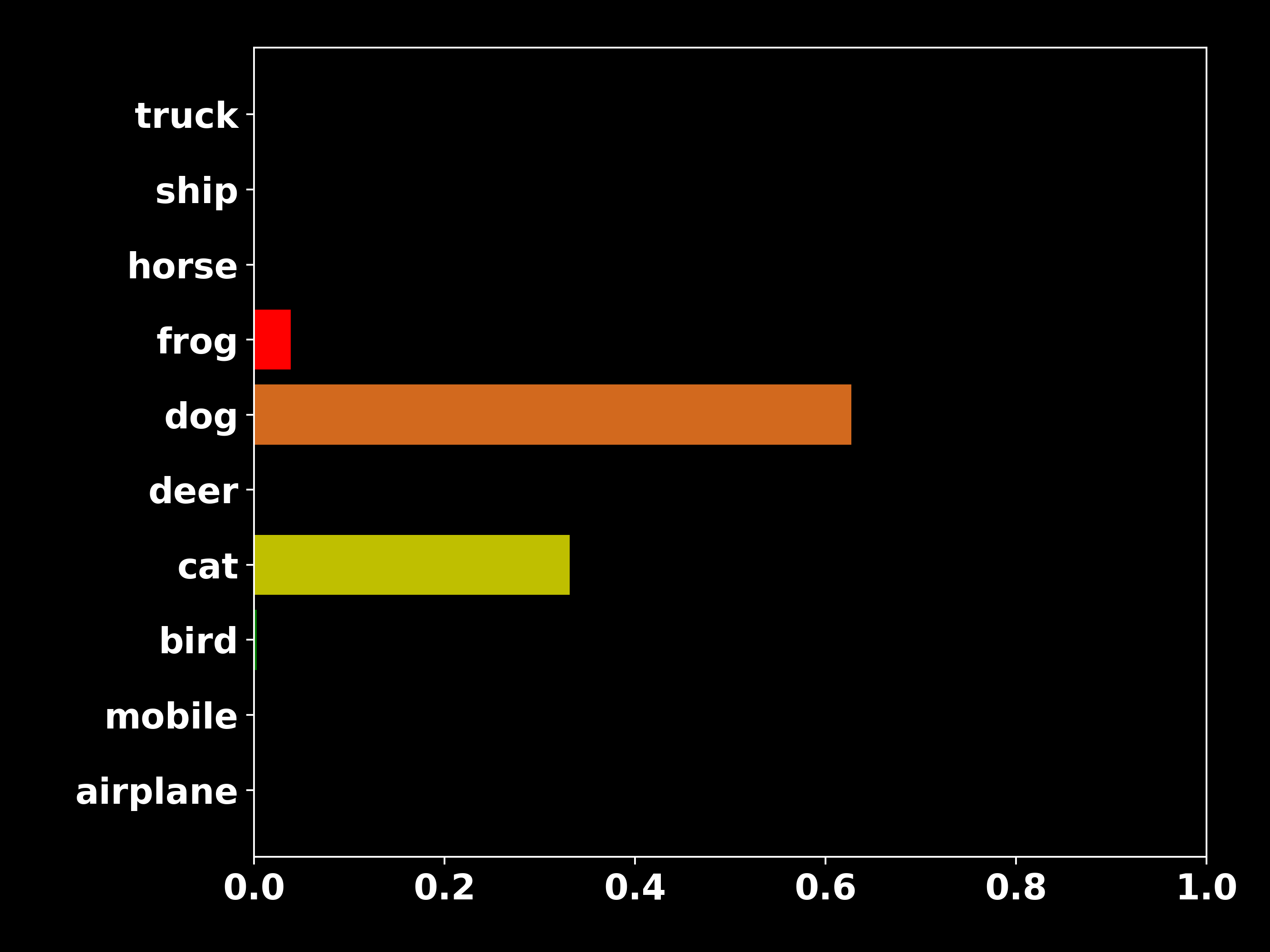}
  \includegraphics[width=.07\linewidth]{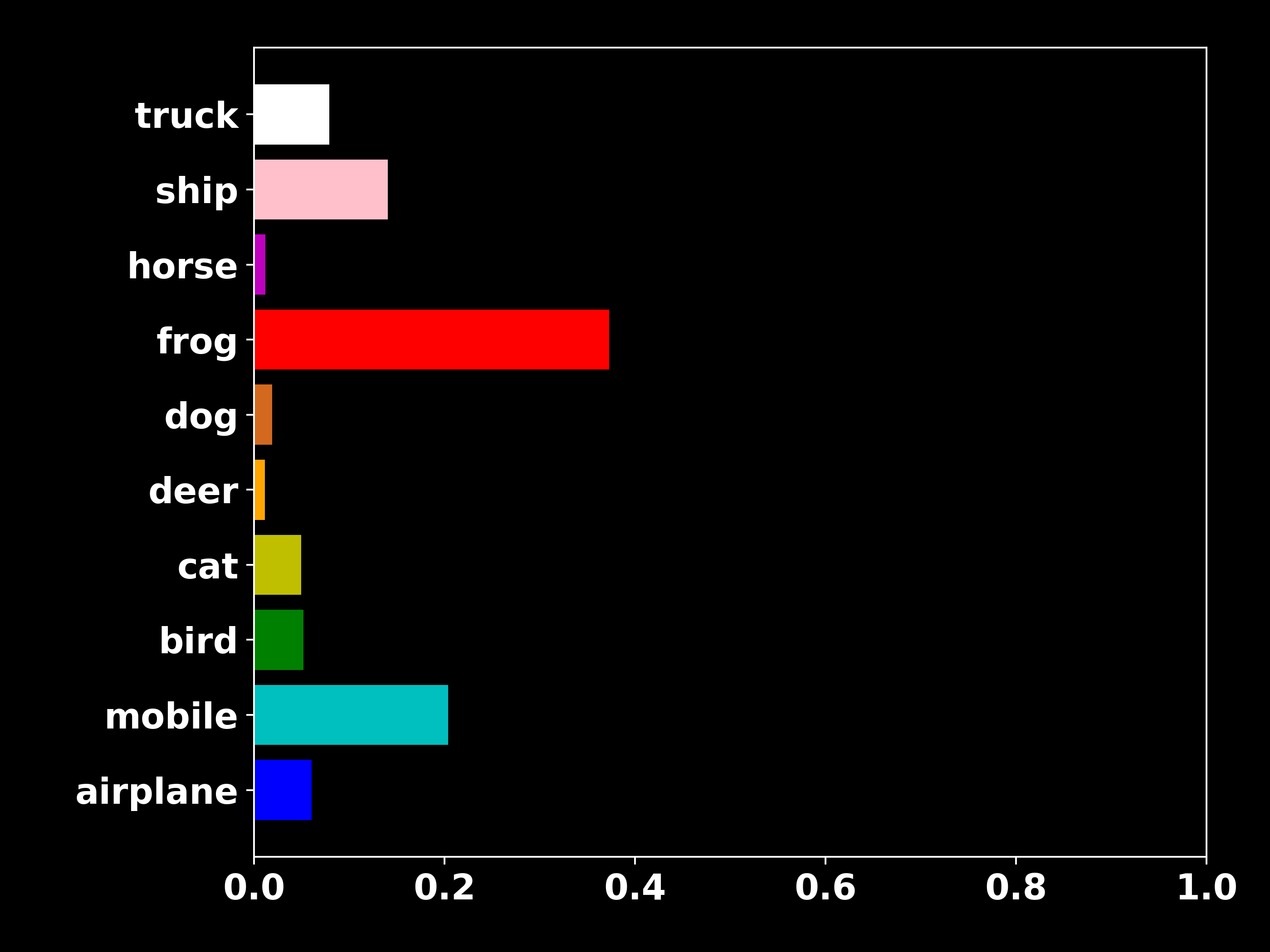}
\end{tabular}
}
\subfloat[A sample of mobile]{
\begin{tabular}[b]{ccc}%
  \includegraphics[width=.07\linewidth]{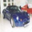}
  \includegraphics[width=.07\linewidth]{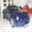}
  \includegraphics[width=.07\linewidth]{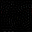} \\
  \includegraphics[width=.07\linewidth]{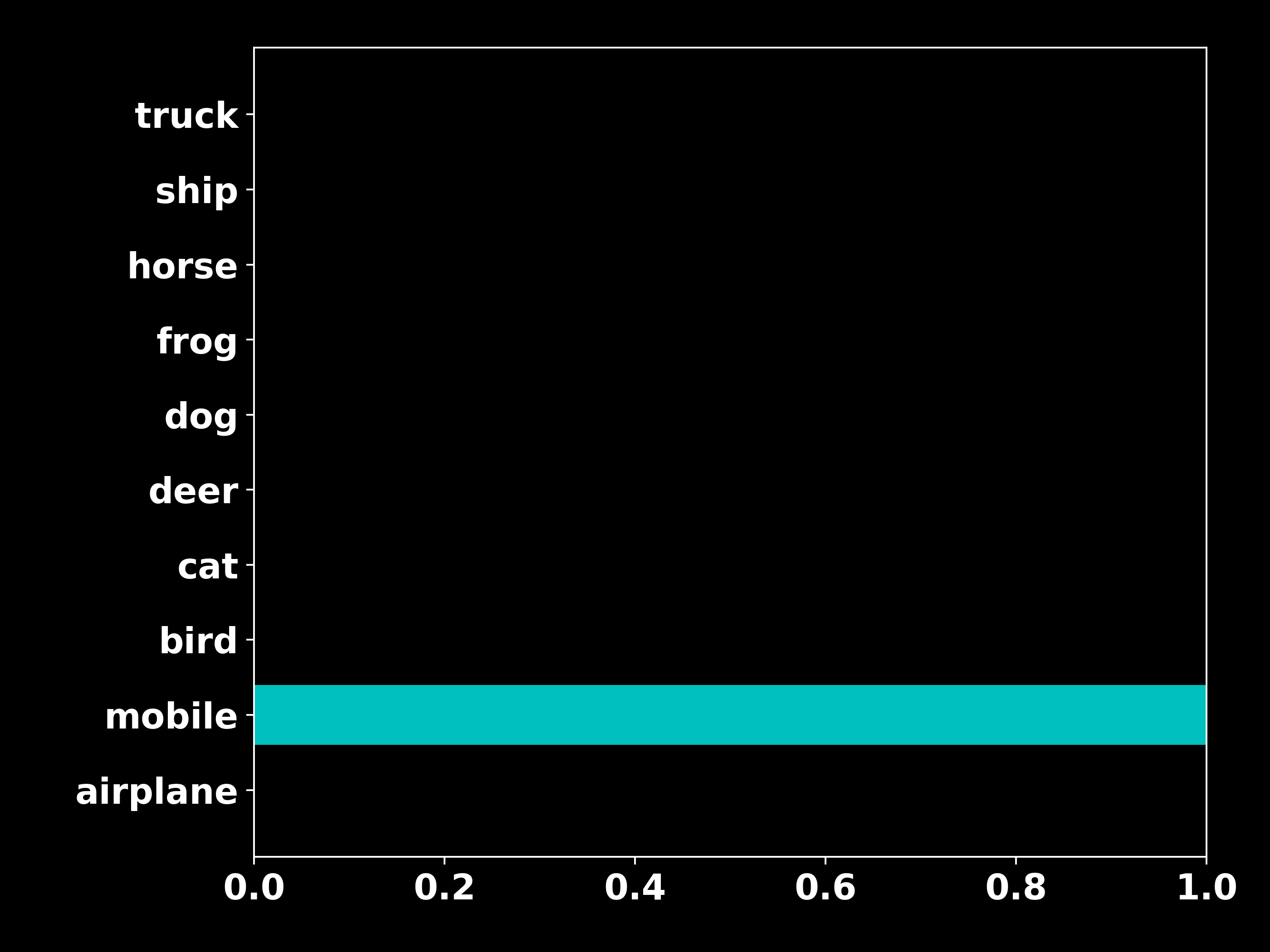}
  \includegraphics[width=.07\linewidth]{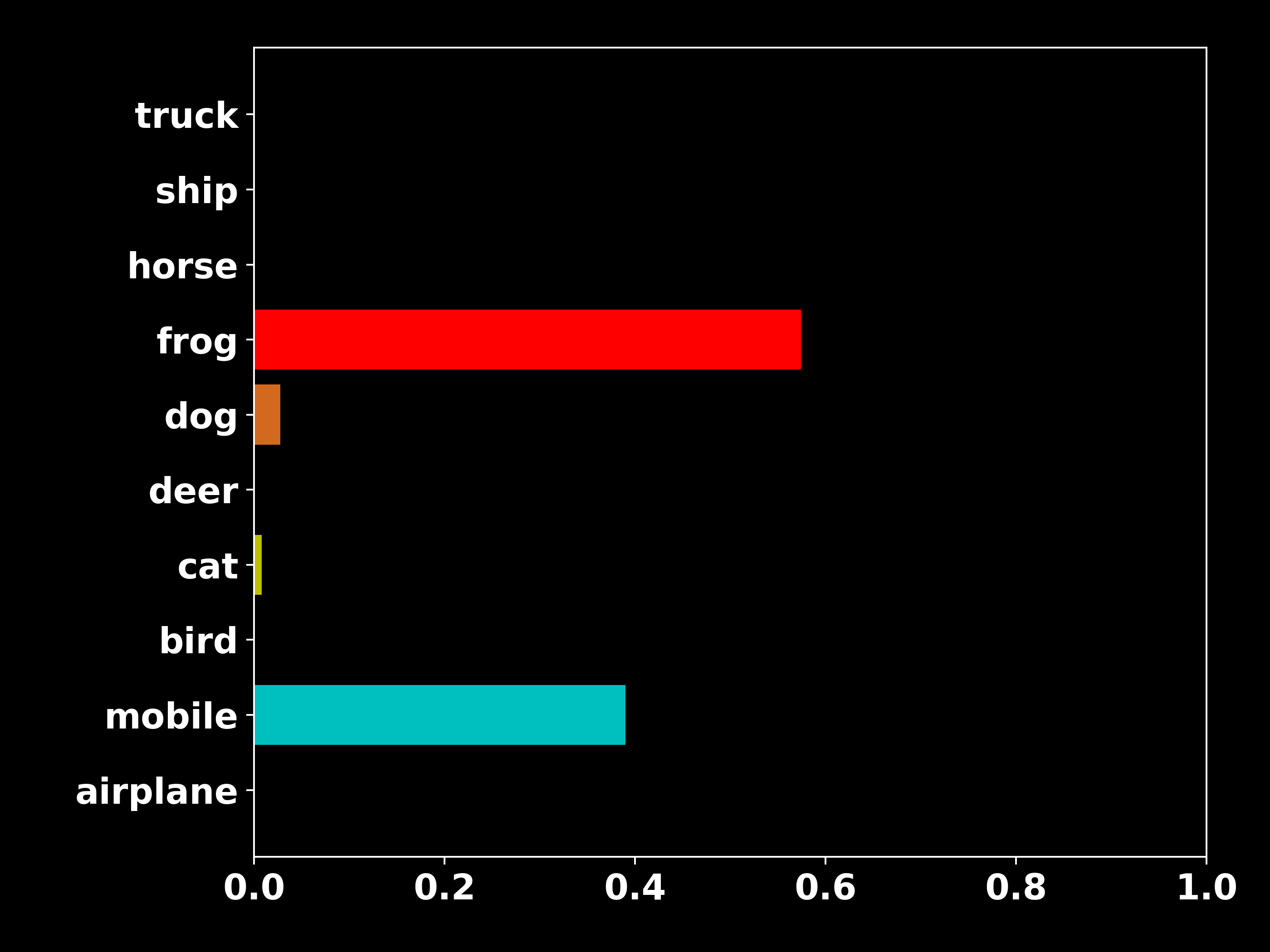}
  \includegraphics[width=.07\linewidth]{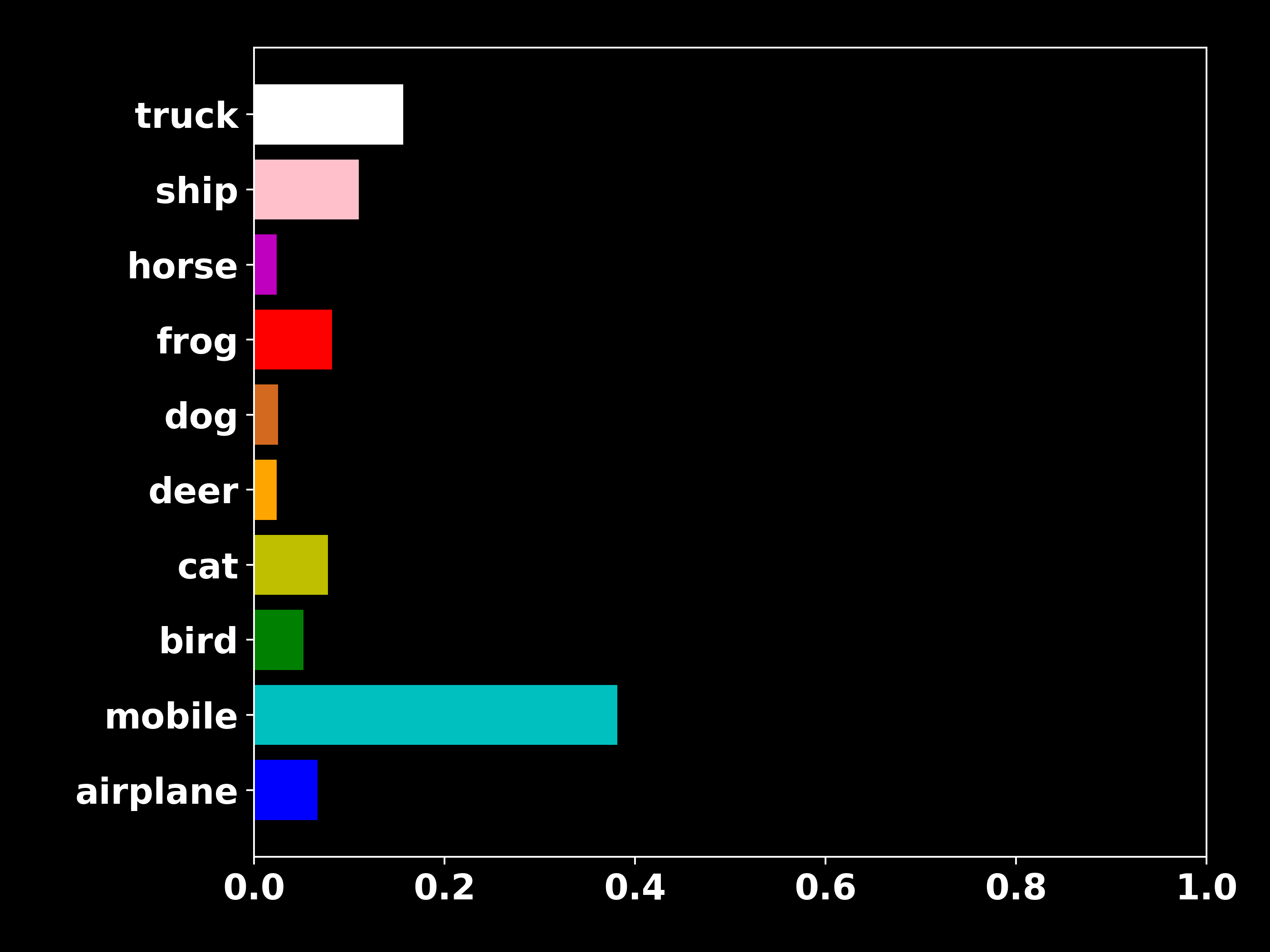}
  \end{tabular}
}
\subfloat[A sample of ship]{
\begin{tabular}[b]{ccc}%
  \includegraphics[width=.07\linewidth]{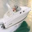}
  \includegraphics[width=.07\linewidth]{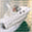}
  \includegraphics[width=.07\linewidth]{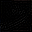} \\
  \includegraphics[width=.07\linewidth]{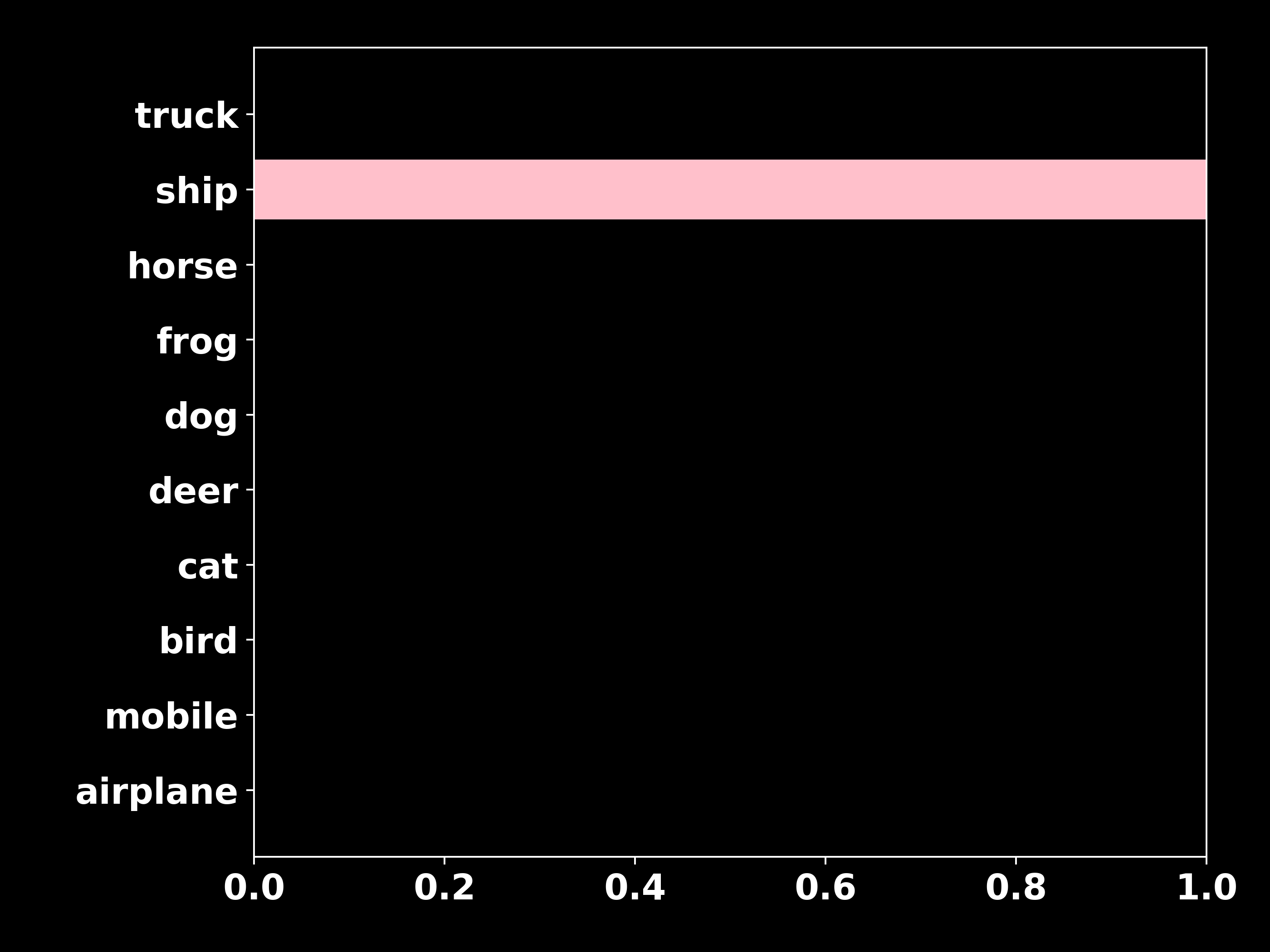}
  \includegraphics[width=.07\linewidth]{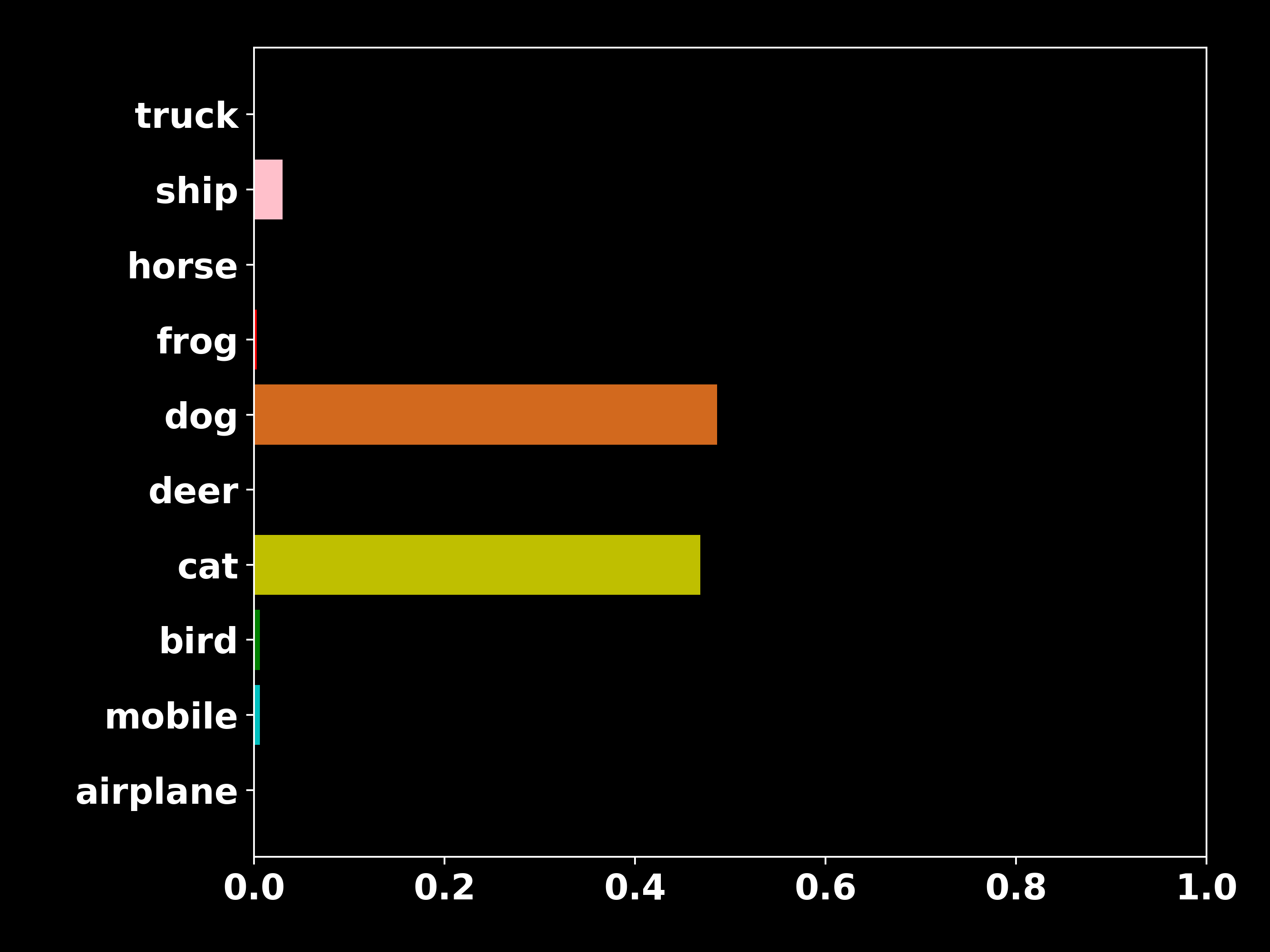}
  \includegraphics[width=.07\linewidth]{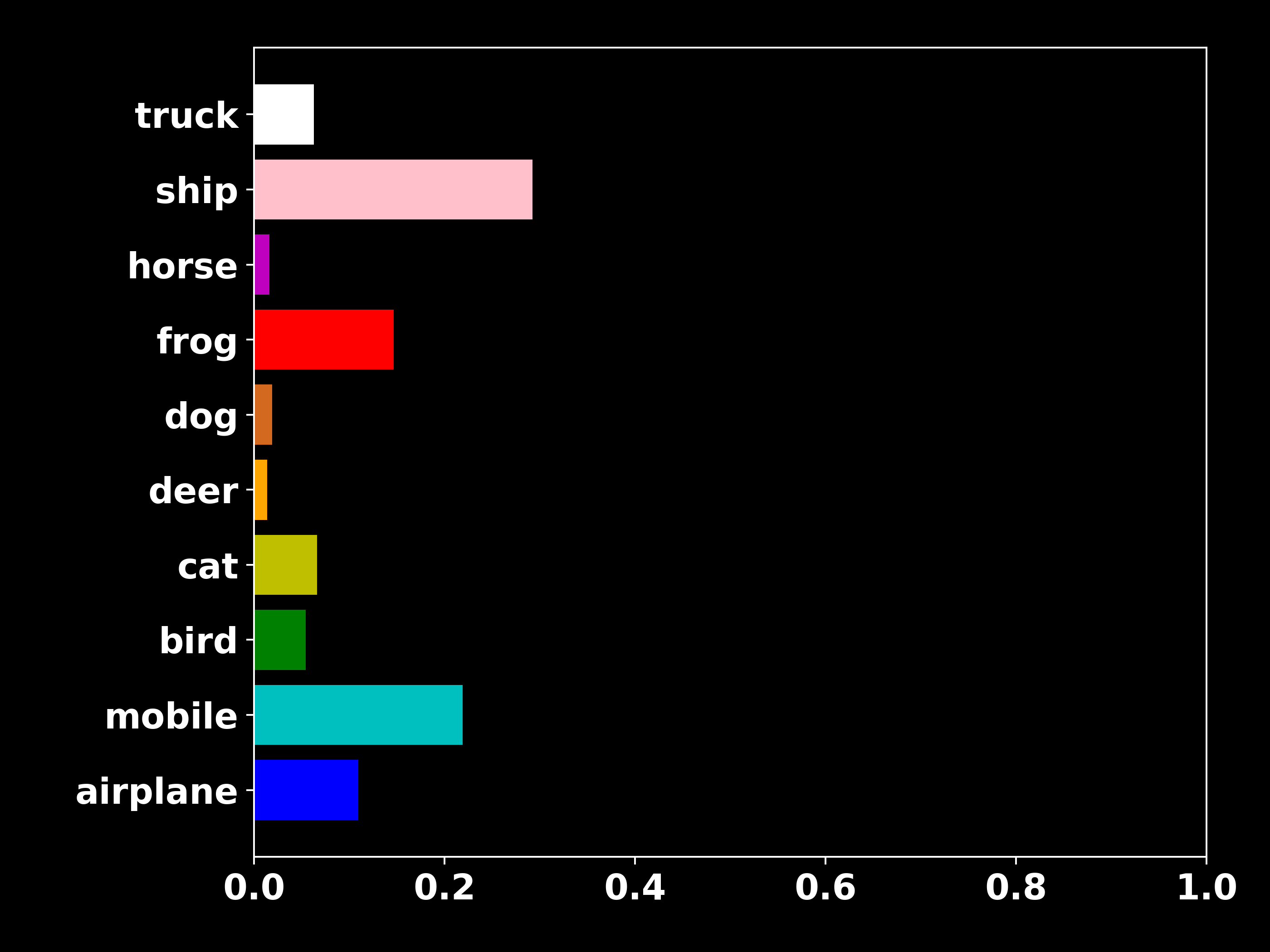}
  \end{tabular}
}
\subfloat[A sample of bird]{
\begin{tabular}[b]{ccc}%
  \includegraphics[width=.07\linewidth]{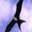}
  \includegraphics[width=.07\linewidth]{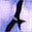}
  \includegraphics[width=.07\linewidth]{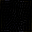} \\
  \includegraphics[width=.07\linewidth]{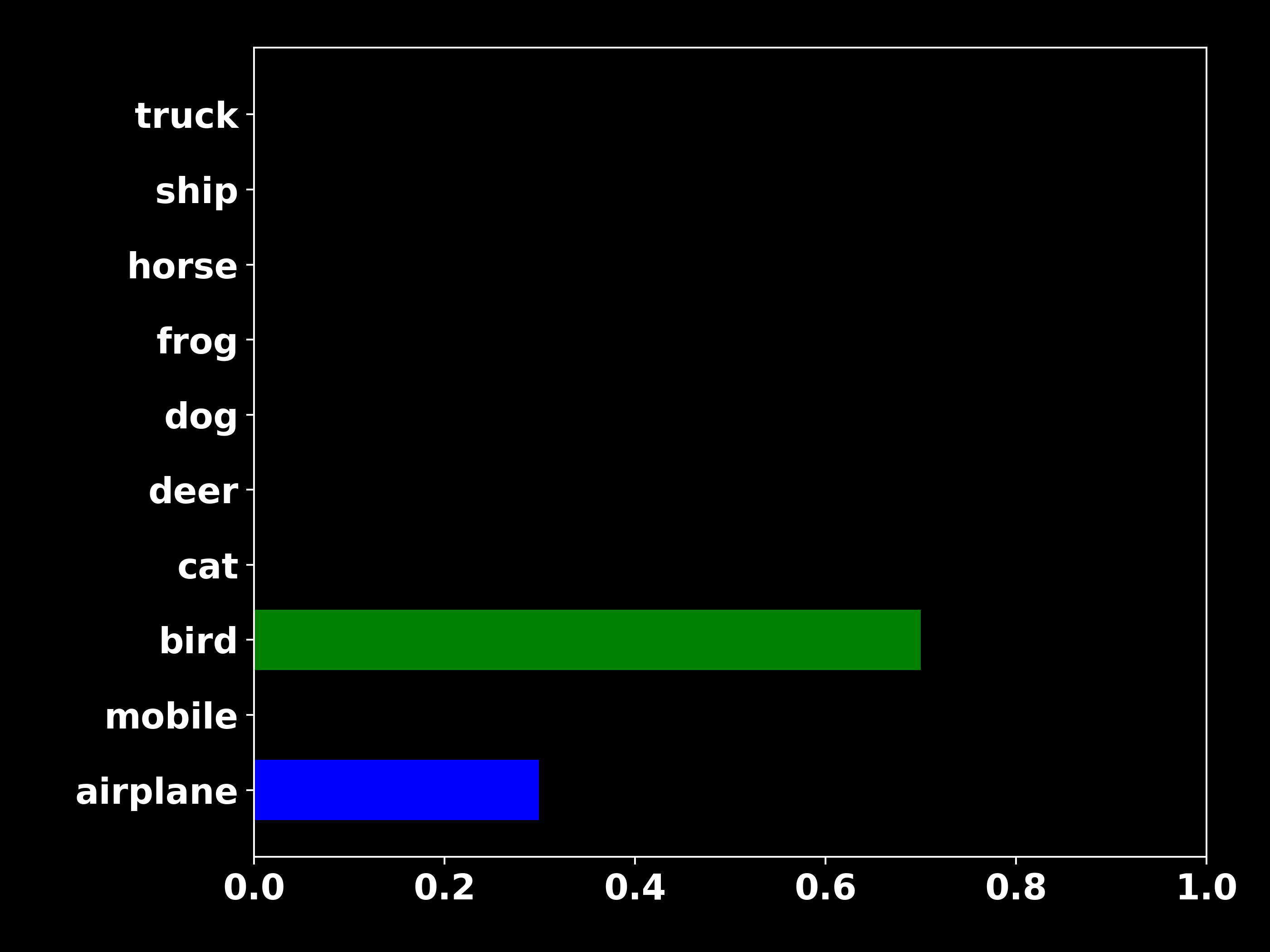}
  \includegraphics[width=.07\linewidth]{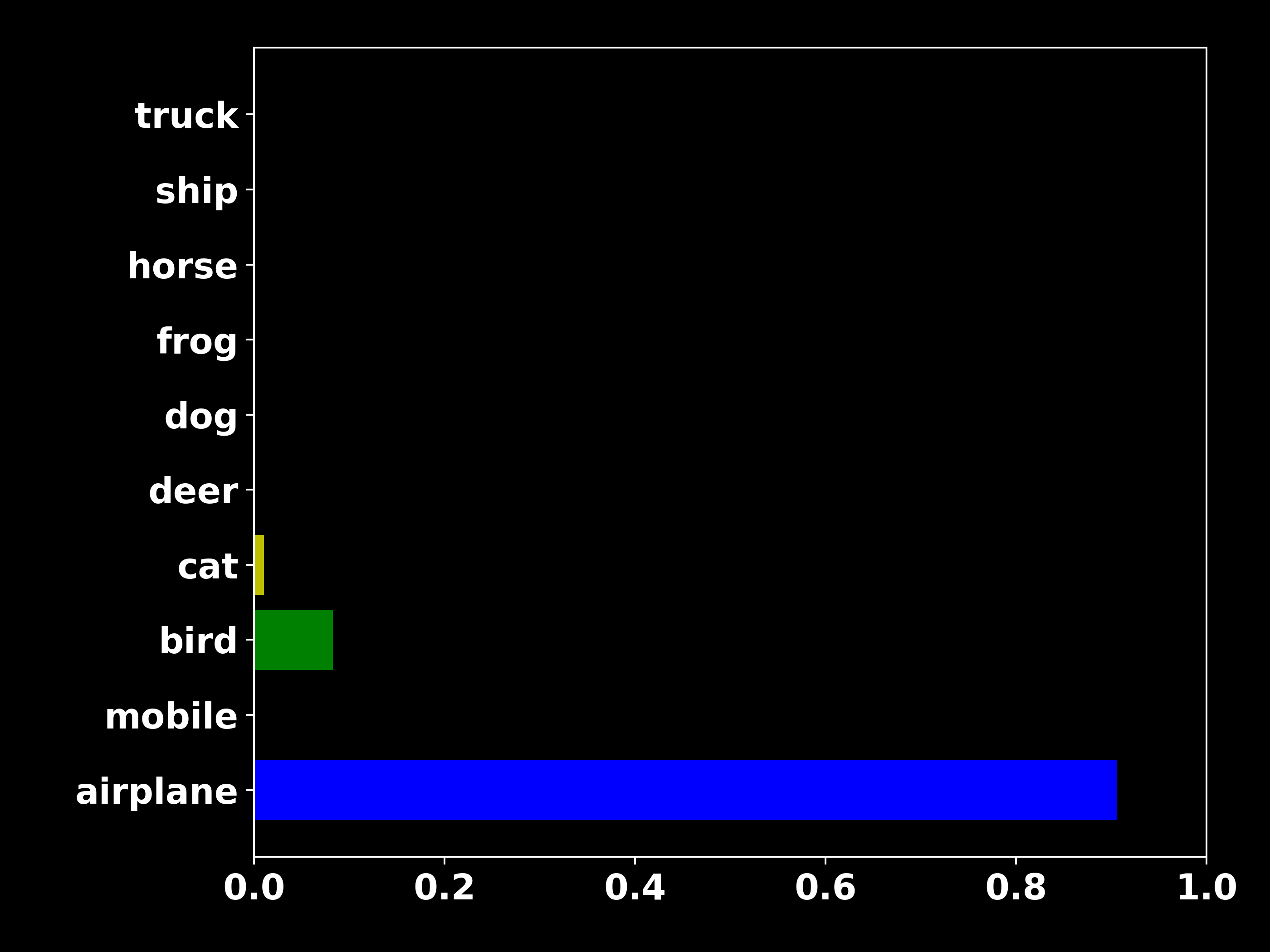}
  \includegraphics[width=.07\linewidth]{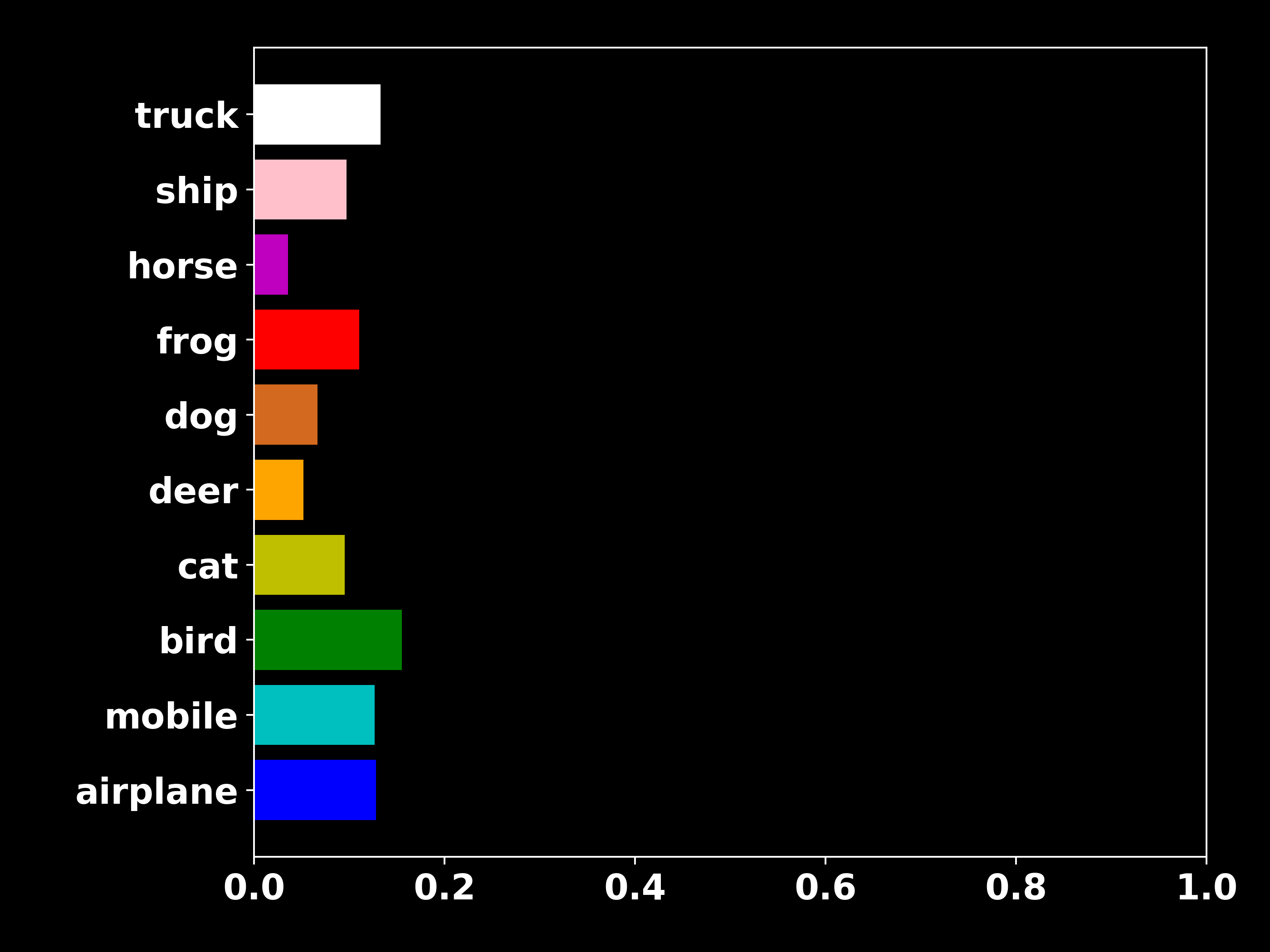}
  \end{tabular}
}
\\
\subfloat[A sample of truck]{
\begin{tabular}[b]{ccc}%
  \includegraphics[width=.07\linewidth]{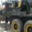}
  \includegraphics[width=.07\linewidth]{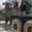}
  \includegraphics[width=.07\linewidth]{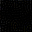} \\
  \includegraphics[width=.07\linewidth]{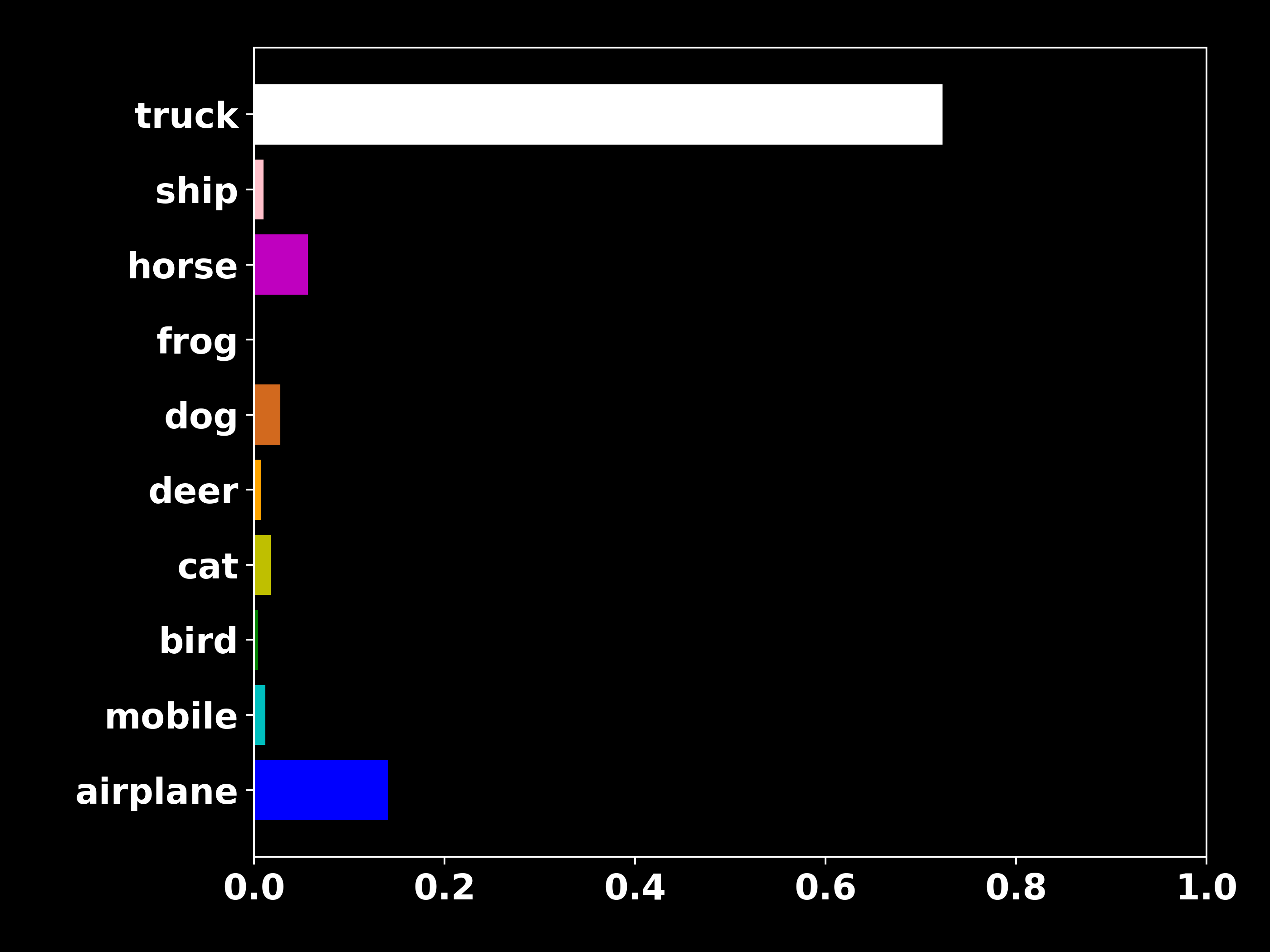}
  \includegraphics[width=.07\linewidth]{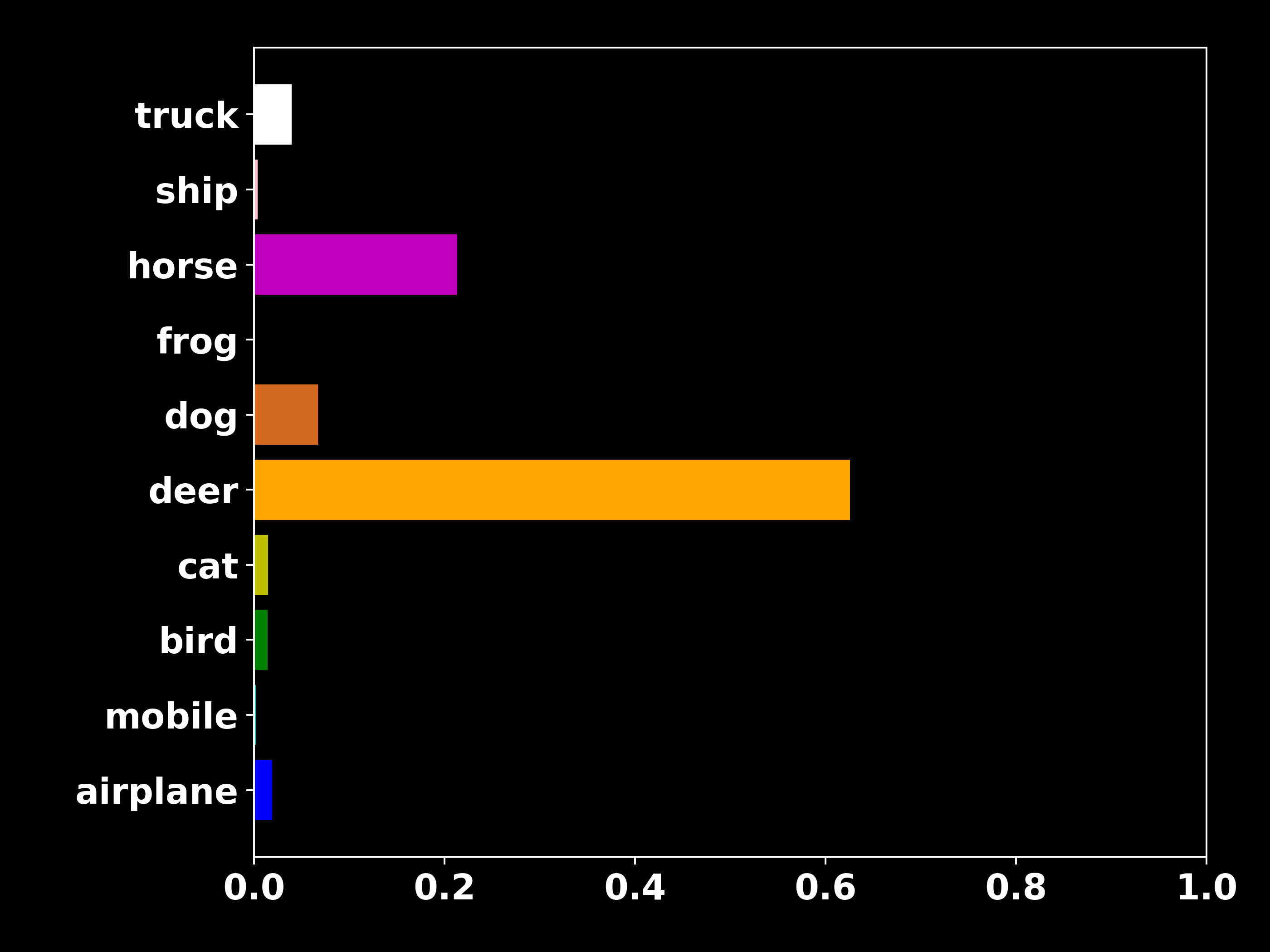}
  \includegraphics[width=.07\linewidth]{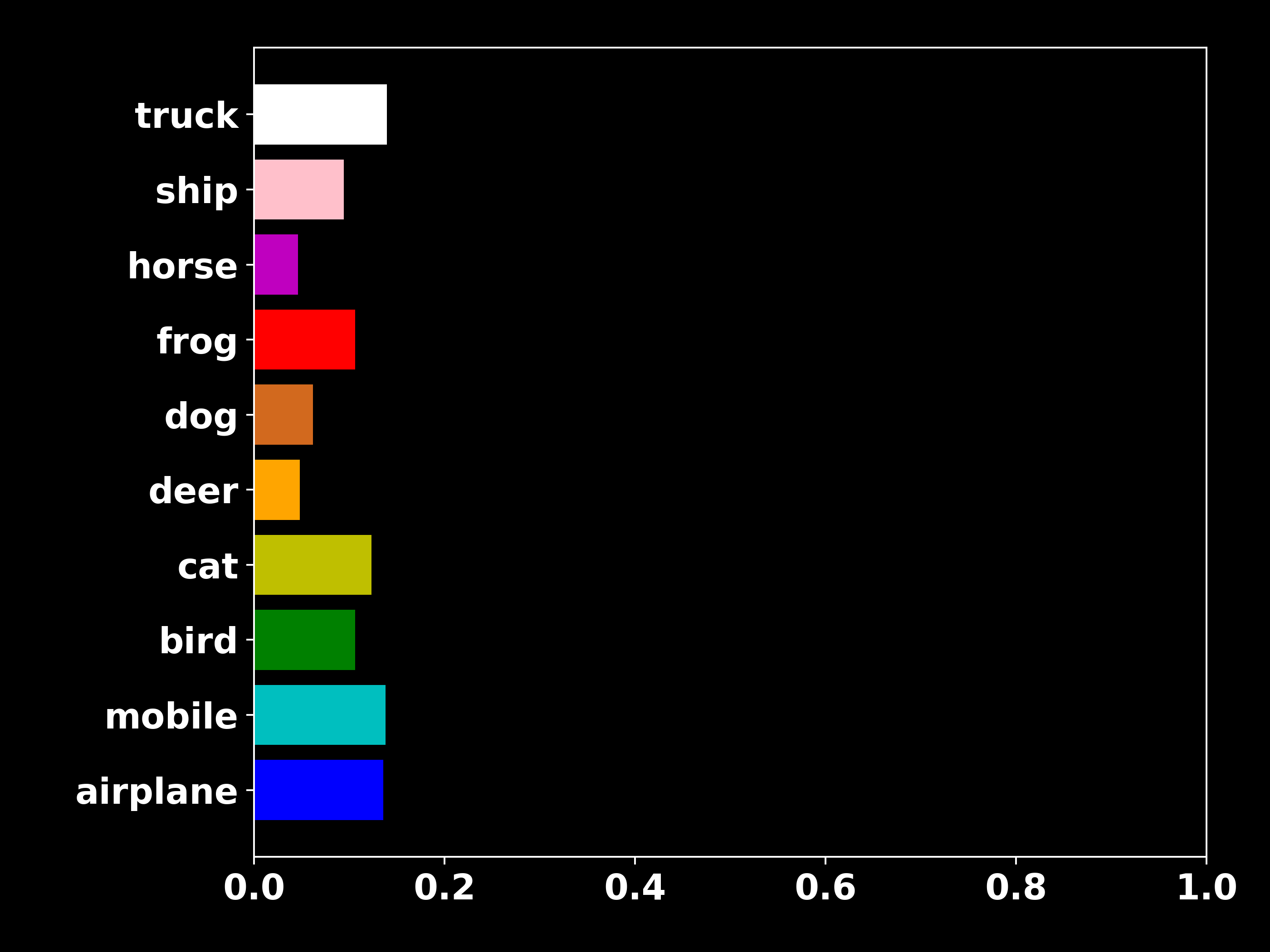}
  \end{tabular}
}
\subfloat[A sample of cat]{
\begin{tabular}[b]{ccc}%
  \includegraphics[width=.07\linewidth]{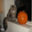}
  \includegraphics[width=.07\linewidth]{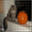}
  \includegraphics[width=.07\linewidth]{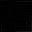} \\
  \includegraphics[width=.07\linewidth]{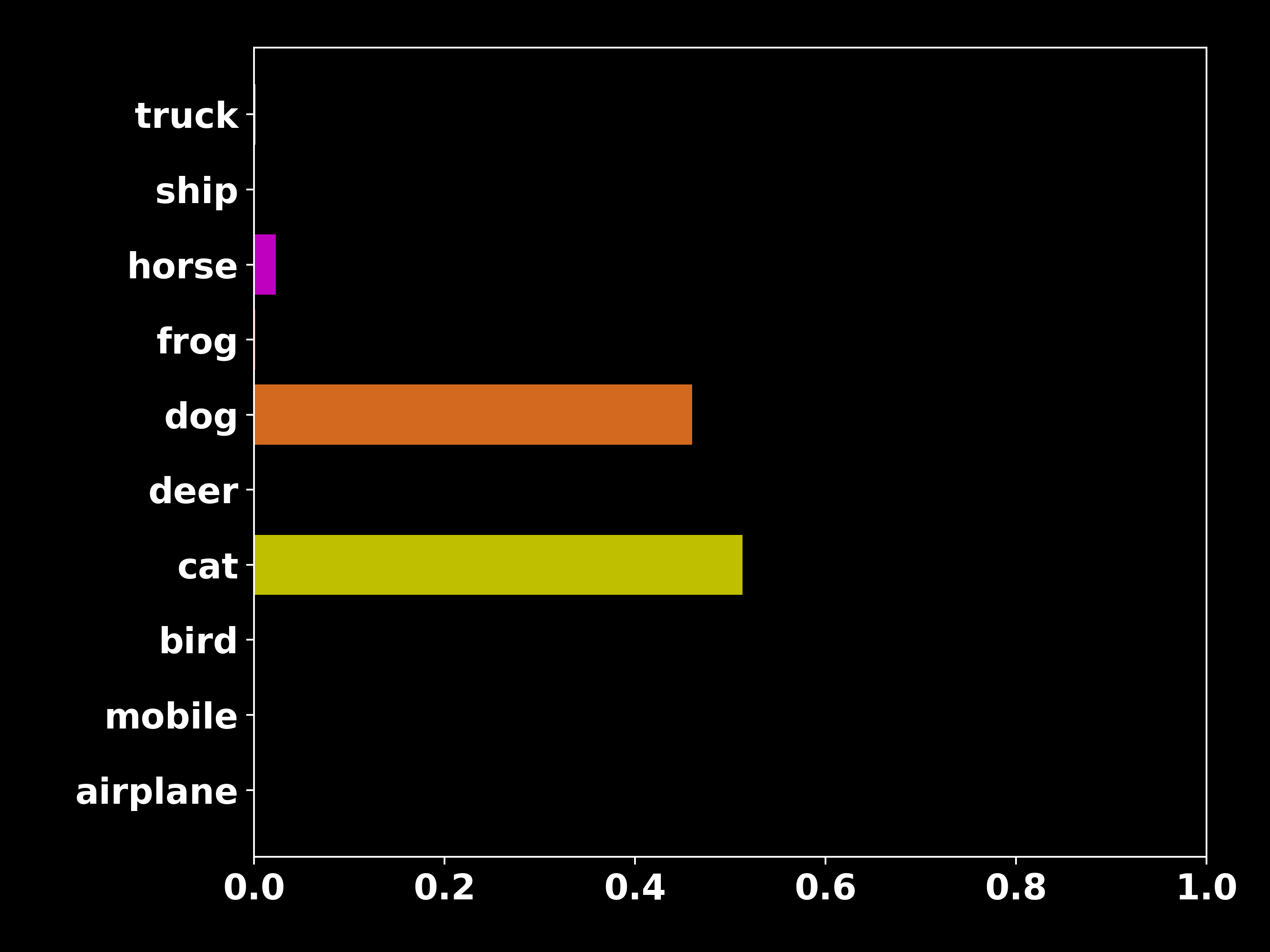}
  \includegraphics[width=.07\linewidth]{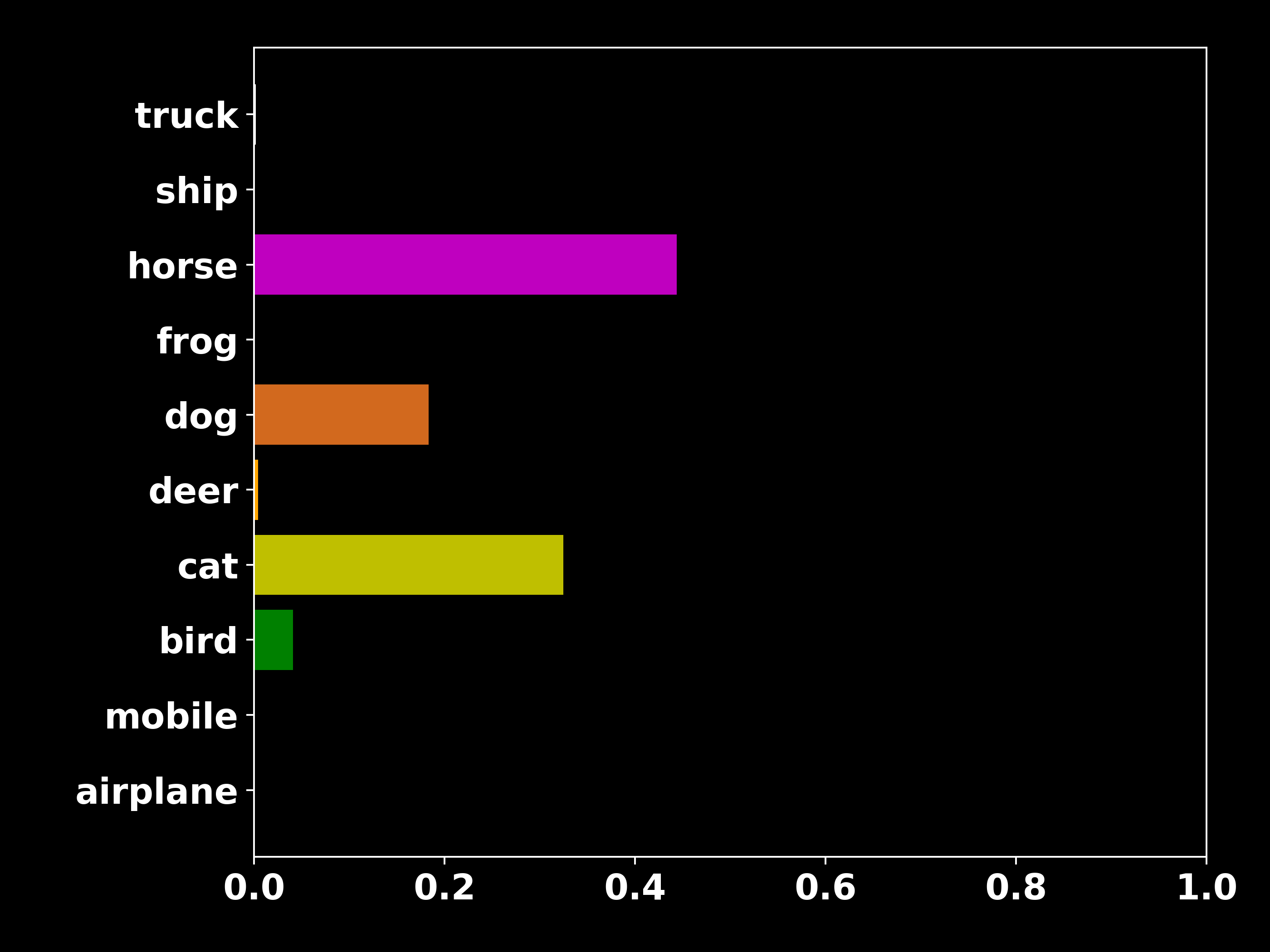}
  \includegraphics[width=.07\linewidth]{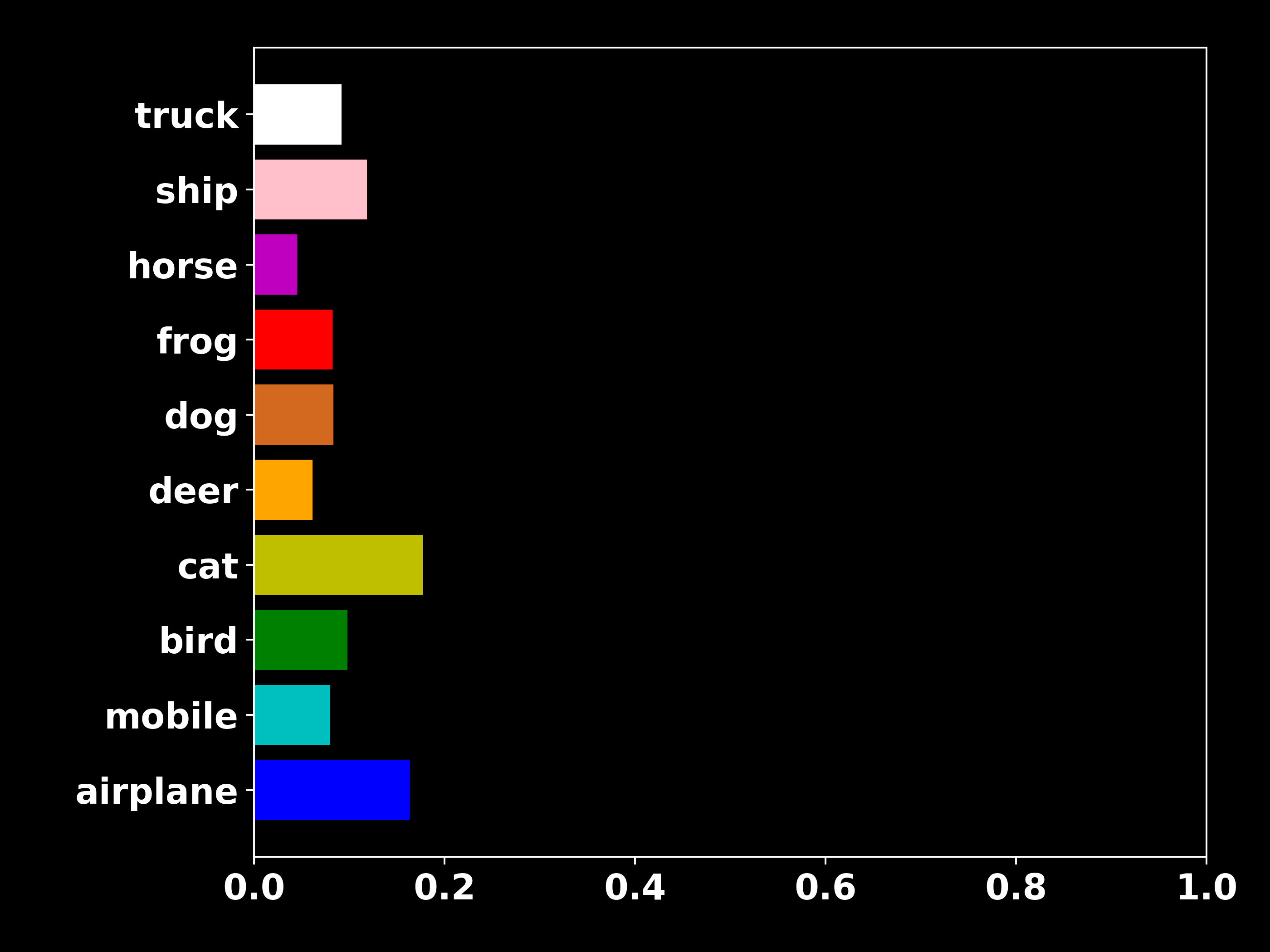}
  \end{tabular}
}
\subfloat[A sample of airplane]{
\begin{tabular}[b]{ccc}%
  \includegraphics[width=.07\linewidth]{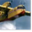}
  \includegraphics[width=.07\linewidth]{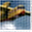}
  \includegraphics[width=.07\linewidth]{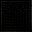} \\
  \includegraphics[width=.07\linewidth]{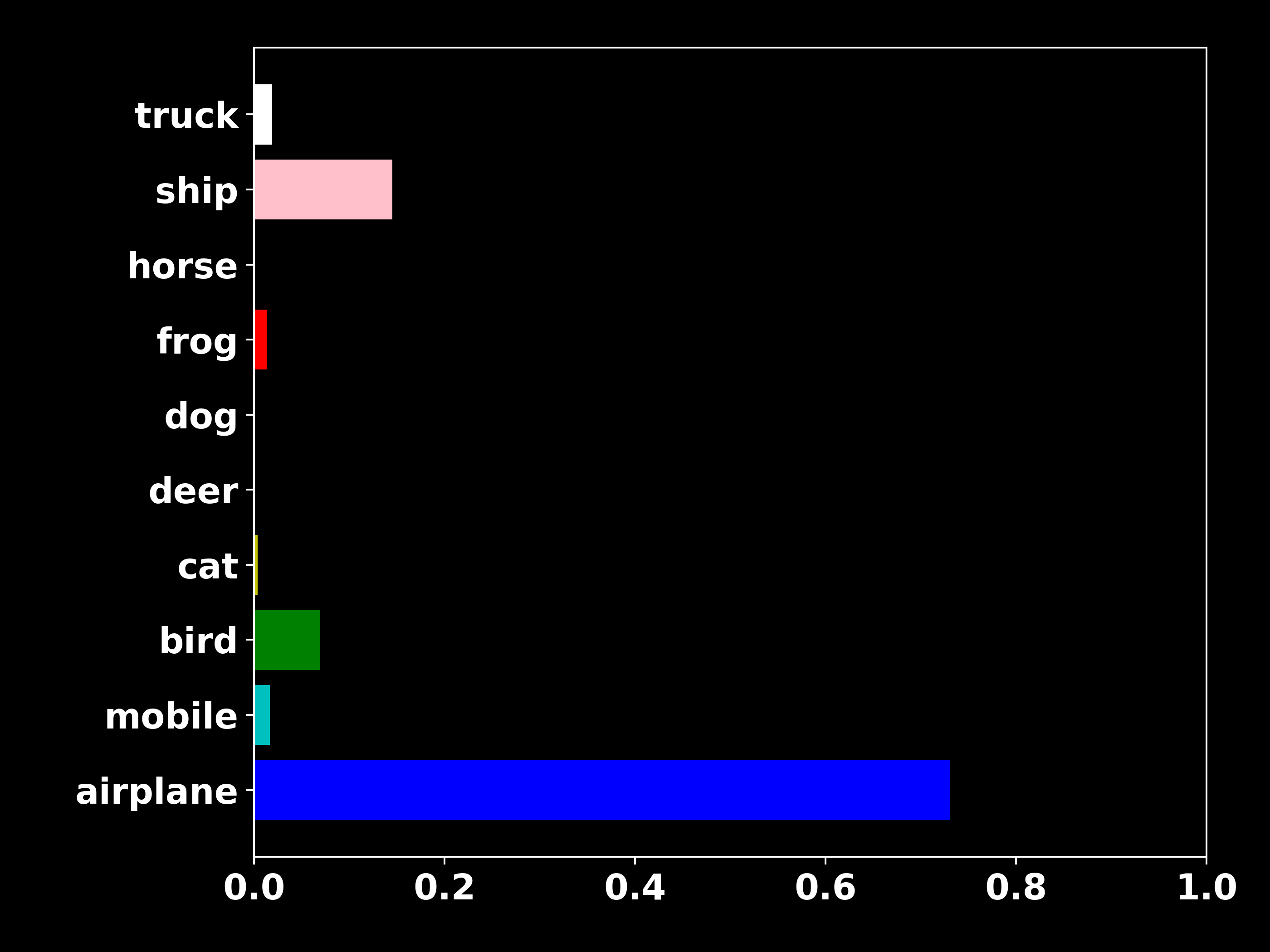}
  \includegraphics[width=.07\linewidth]{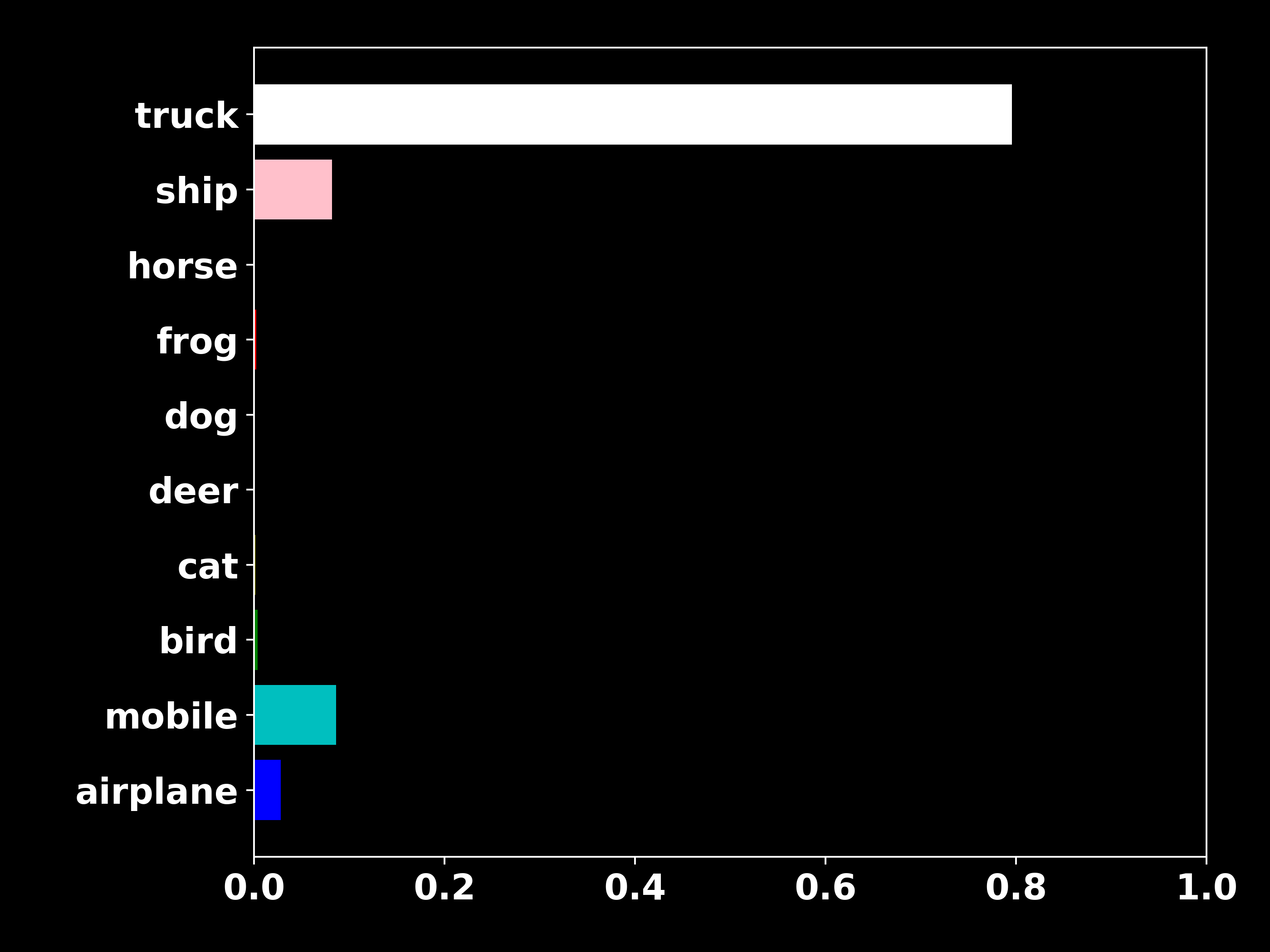}
  \includegraphics[width=.07\linewidth]{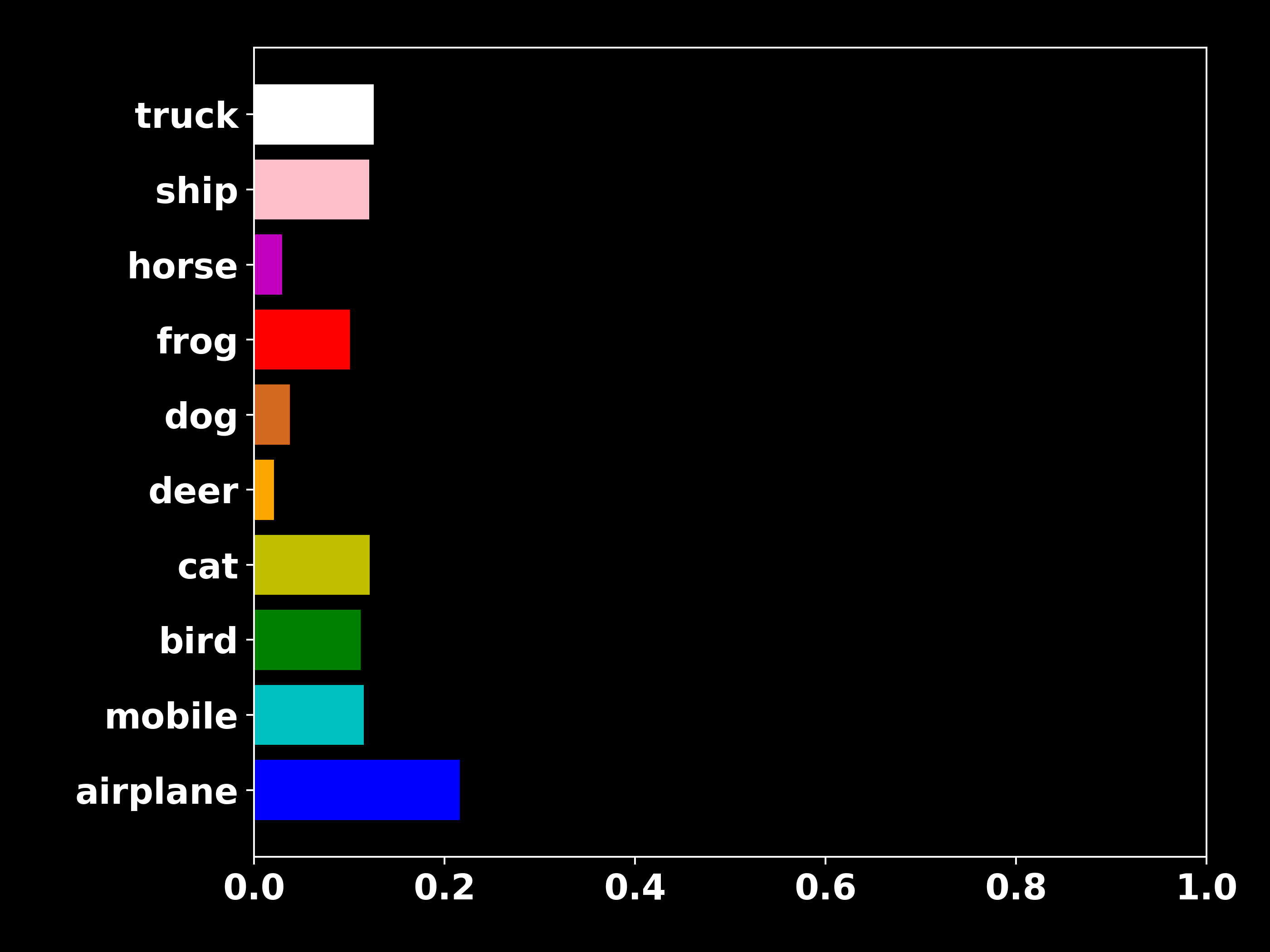}
  \end{tabular}
}
\subfloat[A sample of ship]{
\begin{tabular}[b]{ccc}%
  \includegraphics[width=.07\linewidth]{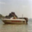}
  \includegraphics[width=.07\linewidth]{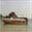}
  \includegraphics[width=.07\linewidth]{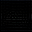} \\
  \includegraphics[width=.07\linewidth]{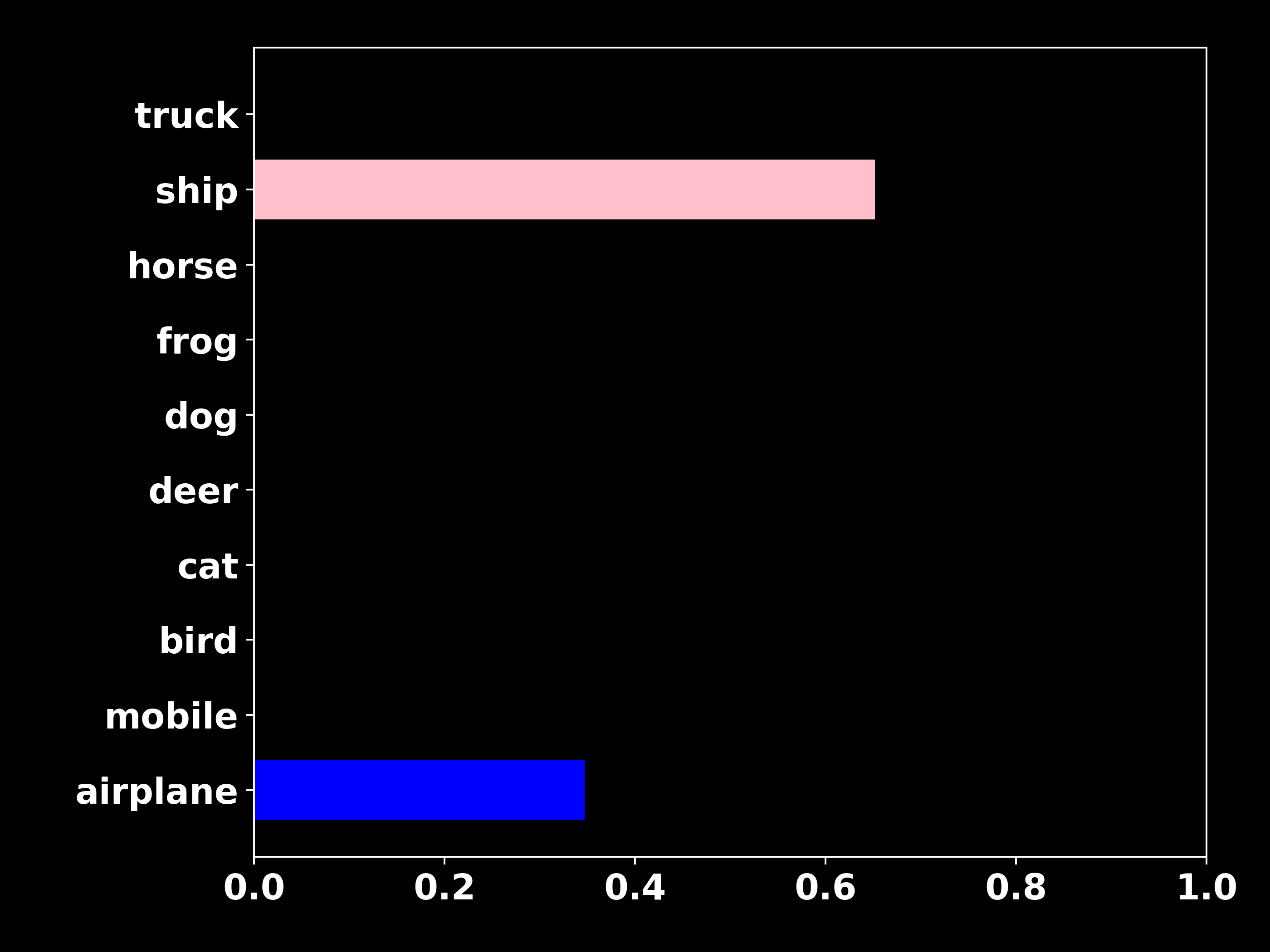}
  \includegraphics[width=.07\linewidth]{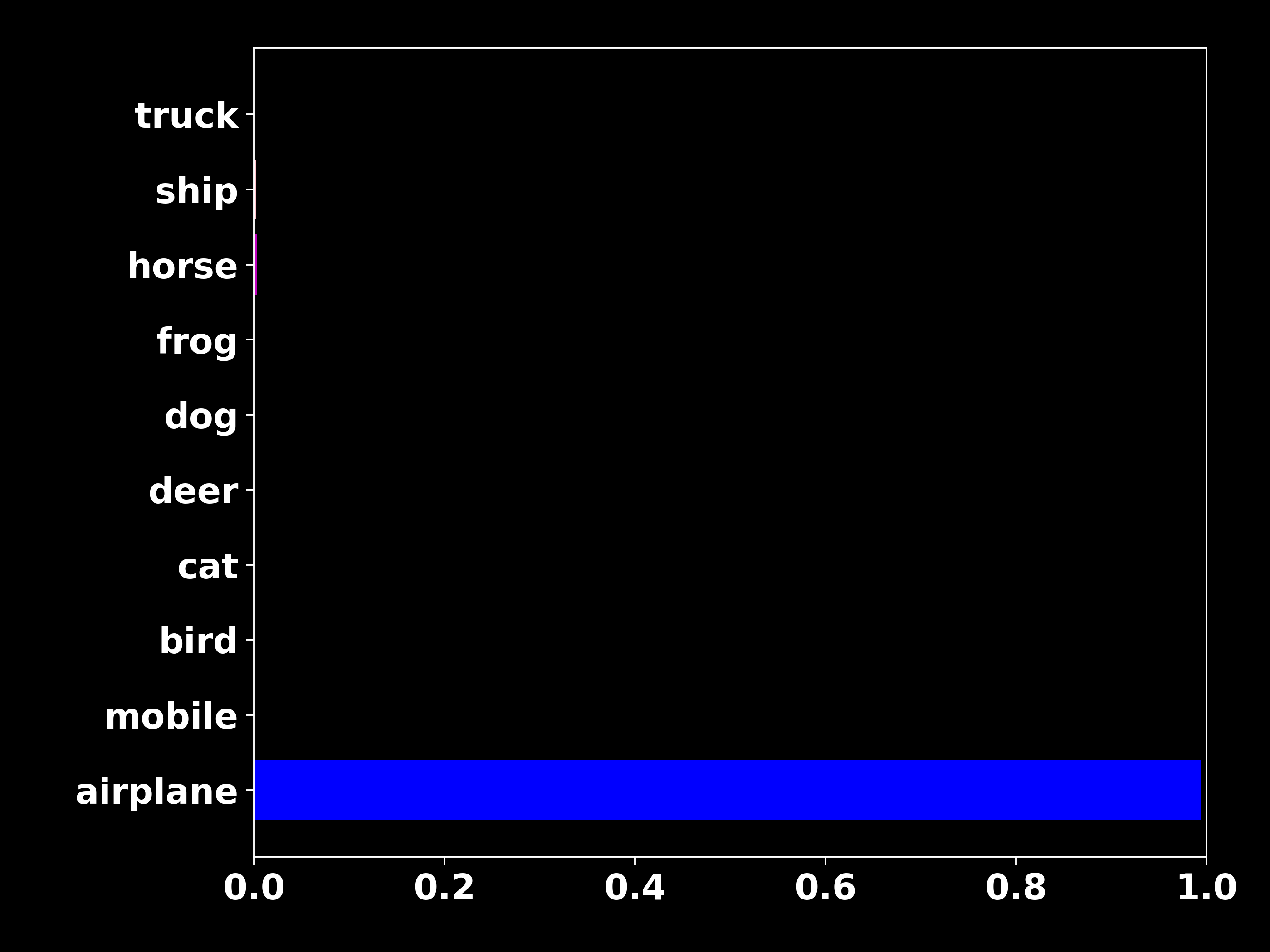}
  \includegraphics[width=.07\linewidth]{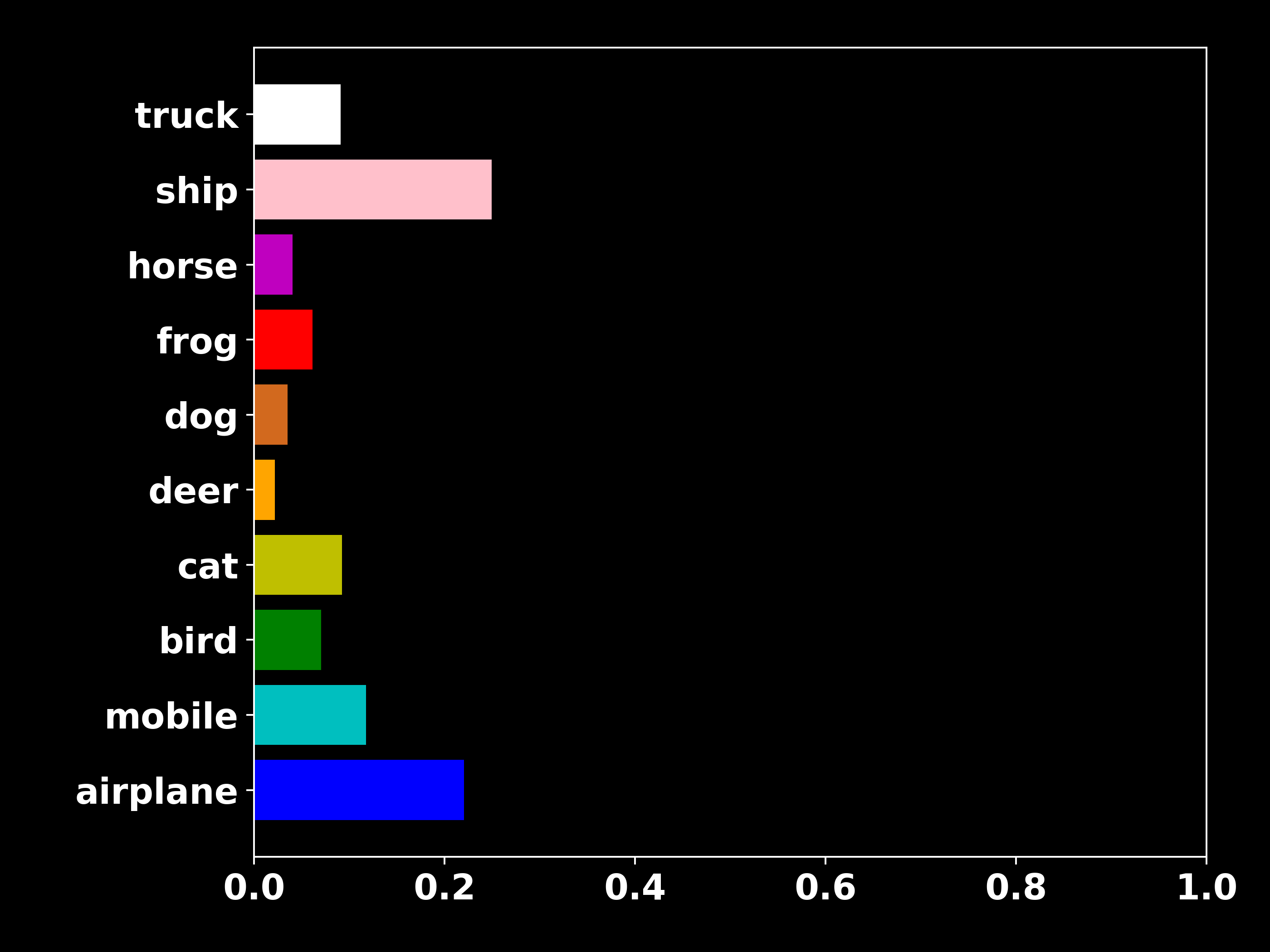}
  \end{tabular}
}
\\
\caption{Eight testing samples selected from CIFAR10 that help explain that CNN can capture the high-frequency image: the model (ResNet18) correctly predicts the original image (1\textsuperscript{st} column in each panel) and the high-frequency reconstructed image (3\textsuperscript{rd} column in each panel), but incorrectly predict the low-frequency reconstructed image (2\textsuperscript{nd} column in each panel). 
The prediction confidences are also shown. 
The frequency components are split with $r=12$. 
Details of the experiment will be introduced later. }
\label{fig:intro}
\end{figure*}

For example, Figure~\ref{fig:intro} shows the prediction results of eight testing samples from CIFAR10 data set, 
together with the prediction results of the high and low-frequency component counterparts. 
For these examples, the prediction outcomes are almost entirely determined by the high-frequency components of the image, which are barely perceivable to human. 
On the other hand, the low-frequency components, which almost look identical to the original image to human, are predicted to something distinctly different by the model. 

Motivated by the above empirical observations, we further investigate the generalization behaviors of CNN and attempt to explain such behaviors via differential responses to the \emph{image frequency spectrum} of the inputs (Remark 1). Our main contributions are summarized as follows: 
\begin{itemize}
    \item We reveal the existing trade-off between CNN's accuracy and robustness by offering examples of how CNN exploits the high-frequency components of images to trade robustness for accuracy (Corollary~\ref{theorem:main}). 
    \item With image frequency spectrum as a tool, we offer hypothesis to explain several generalization behaviors of CNN, especially the capacity in memorizing label-shuffled data. 
    \item We propose defense methods that can help improving the adversarial robustness of CNN towards simple attacks without training or fine-tuning the model. 
\end{itemize}

The remainder of the paper is organized as follows. 
In Section~\ref{sec:related}, we first introduce related discussions. 
In Section~\ref{sec:method}, we will present our main contributions, 
including a formal discussion on that CNN can exploit high-frequency components, 
which naturally leads to the trade-off between adversarial robustness and accuracy.
Further, 
in Section~\ref{sec:rethinking}-\ref{sec:adversarial},
we set forth to investigate multiple generalization behaviors of CNN, 
including the paradox related to capacity of memorizing label-shuffled data (\cref{sec:rethinking}), 
the performance boost introduced by heuristics such as Mixup and BatchNorm (\cref{sec:heuristics}), 
and the adversarial vulnerability (\cref{sec:adversarial}).  
We also attempt to investigate tasks beyond image classification in Section~\ref{sec:beyond}. 
Finally, we will briefly discuss some related topics in Section~\ref{sec:discussion} before we conclude the paper in Section~\ref{sec:con}.


\section{Related Work}
\label{sec:related}
The remarkable success of deep learning has attracted a torrent of theoretical work devoted to explaining the generalization mystery of CNN. 

For example, ever since Zhang \textit{et al.} \cite{zhang2016understanding} demonstrated the effective capacity of several successful neural network architectures is large enough to memorize random labels, the community sees a prosperity of many discussions about this apparent ''paradox'' \cite{wu2017towards, dinh2017sharp, dziugaite2017computing, dinh2017sharp, chen2018closing}. 
Arpit \textit{et al.} \cite{arpit2017closer} demonstrated that effective capacity are unlikely to explain the generalization performance of gradient-based-methods trained deep networks due to the training data largely determine memorization. 
Kruger \textit{et al.}\cite{krueger2017deep} empirically argues by showing largest Hessian eigenvalue increased when training on random labels in deep networks.

The concept of adversarial example \cite{szegedy2013intriguing,goodfellow2015explaining} has become another intriguing direction relating to the behavior of neural networks. 
Along this line, 
researchers invented powerful methods such as FGSM~\cite{goodfellow2015explaining}, PGD~\cite{madry2017towards}, and many others \cite{xiao2018generating,carlini2017towards,su2019one,kurakin2016adversarial,chen2017targeted} to deceive the models. This is known as \emph{attack methods}.
In order to defend the model against the deception, another group of researchers proposed a wide range of methods (known as \emph{defense methods})~\cite{akhtar2018defense,lee2017generative,meng2017magnet,metzen2017detecting,he2017adversarial}. 
These are but a few highlights among a long history of proposed attack and defense methods. 
One can refer to comprehensive reviews for detailed discussions \cite{Akhtar_2018,chakraborty2018adversarial}

However, while improving robustness, these methods may see a slight drop of prediction accuracy, 
which leads to another thread of discussion in the trade-off between robustness and accuracy. The empirical results in \cite{rozsa2016accuracy} demonstrated that more accurate model tend to be more robust over generated adversarial examples. While \cite{hendrycks2019benchmarking} argued that the seemingly increased robustness are mostly due to the increased accuracy, and more accurate models (\textit{e.g.}, VGG, ResNet) are actually less robust than AlexNet. 
Theoretical discussions have also been offered \cite{tsipras2018robustness, zhang2019theoretically}, which also inspires new defense methods \cite{zhang2019theoretically}. 

\section{High-frequency Components \& CNN's Generalization}
\label{sec:method}
We first set up the basic notations used in this paper: 
$\langle \mathbf{x}, \mathbf{y} \rangle$ denotes a data sample (the image and the corresponding label).
$f(\cdot;\theta)$ denotes a convolutional neural network whose parameters are denoted as $\theta$. 
We use $\mathcal{H}$ to denote a human model, and as a result, $f(\cdot;\mathcal{H})$ denotes how human will classify the data $\cdot$.
$l(\cdot, \cdot)$ denotes a generic loss function (\textit{e.g.}, cross entropy loss).  
$\alpha(\cdot, \cdot)$ denotes a function evaluating prediction accuracy (for every sample, this function yields $1.0$ if the sample is correctly classified, $0.0$ otherwise). 
$d(\cdot, \cdot)$ denotes a function evaluating the distance between two vectors. 
$\mathcal{F}(\cdot)$ denotes the Fourier transform; thus, $\mathcal{F}^{-1}(\cdot)$ denotes the inverse Fourier transform. 
We use $\mathbf{z}$ to denote the frequency component of a sample. 
Therefore, we have $\mathbf{z} = \mathcal{F}(\mathbf{x})$ and $\mathbf{x} = \mathcal{F}^{-1}(\mathbf{z})$. 

Notice that Fourier transform or its inverse may introduce complex numbers.  
In this paper, we simply discard the imaginary part of the results of $\mathcal{F}^{-1}(\cdot)$ to make sure 
the resulting image can be fed into CNN as usual. 



\subsection{CNN Exploit High-frequency Components}

We decompose the raw data $\mathbf{x} = \{\mathbf{x}_l, \mathbf{x}_h\}$, where $\mathbf{x}_l$ and $\mathbf{x}_h$ denote the low-frequency component (shortened as \lfc{}) and high-frequency component (shortened as \hfc{}) of $\mathbf{x}$. 
We have the following four equations: 
\begin{align*}
    \mathbf{z} = \mathcal{F}(\mathbf{x}), \quad \quad
    \mathbf{z}_l, \mathbf{z}_h =t(\mathbf{z}; r), \\
    \mathbf{x}_l = \mathcal{F}^{-1}(\mathbf{z}_l), \quad \quad
    \mathbf{x}_h = \mathcal{F}^{-1}(\mathbf{z}_h),
\end{align*}
where $t(\cdot; r)$ denotes a thresholding function that separates the low and high frequency components from $\mathbf{z}$ according to a hyperparameter, radius $r$. 

To define $t(\cdot; r)$ formally, we first consider a grayscale (one channel) image of size $n \times n$ with $\mathcal{N}$ possible pixel values (in other words, $\mathbf{x} \in \mathcal{N}^{n \times n}$), then we have $\mathbf{z} \in \mathcal{C}^{n \times n}$, where $\mathcal{C}$ denotes the complex number. 
We use $\mathbf{z}(i,j)$ to index the value of $\mathbf{z}$ at position $(i,j)$, and we use $c_i, c_j$ to denote the centroid. 
We have the equation $\mathbf{z}_l, \mathbf{z}_h =t(\mathbf{z}; r)$ formally defined as:
\begin{align*}
    \mathbf{z}_l(i,j) = \begin{cases} \mathbf{z}(i,j), \,&\textnormal{if} \,  d((i,j),(c_i,c_j)) \leq r \\
    0, \, &\textnormal{otherwise}
    \end{cases},
    \\
    \mathbf{z}_h(i,j) = \begin{cases} 0, \, &\textnormal{if} \,d((i,j),(c_i,c_j)) \leq r \\
    \mathbf{z}(i,j), \, &\textnormal{otherwise}
    \end{cases}
\end{align*}

We consider $d(\cdot, \cdot)$ in $t(\cdot; r)$ as the Euclidean distance in this paper. 
If $\mathbf{x}$ has more than one channel, then the procedure operates on every channel of pixels independently.

\begin{remark}
\label{remark:main}
With an assumption (referred to as A1) that presumes ``only $\mathbf{x}_l$ is perceivable to human, but both $\mathbf{x}_l$ and $\mathbf{x}_h$ are perceivable to a CNN,''
we have:
\begin{align*}
    \mathbf{y} := f(\mathbf{x};\mathcal{H}) = f(\mathbf{x}_l;\mathcal{H}),
\end{align*}
but when a CNN is trained with
\begin{align*}
    \argmin_\theta l(f(\mathbf{x};\theta), \mathbf{y}), 
\end{align*}
which is equivalent to
\begin{align*}
    \argmin_\theta l(f(\{\mathbf{x}_l, \mathbf{x}_h\};\theta), \mathbf{y}),
\end{align*}
CNN may learn to exploit $\mathbf{x}_h$ to minimize the loss. 
As a result, CNN's generalization behavior appears unintuitive to a human. \qed
\end{remark}

Notice that ``CNN may learn to exploit $\mathbf{x}_h$'' differs from ``CNN overfit'' because $\mathbf{x}_h$ can contain more information than sample-specific idiosyncrasy, and these more information can be generalizable across training, validation, and testing sets, but are just imperceptible to a human.

As Assumption A1 has been demonstrated to hold in some cases (\textit{e.g.}, in Figure~\ref{fig:intro}), 
we believe Remark~\ref{remark:main} can serve as one of the explanations to CNN's generalization behavior. 
For example, the adversarial examples \cite{szegedy2013intriguing,goodfellow2015explaining} 
can be generated by perturbing $\mathbf{x}_h$; 
the capacity of CNN in reducing training error to zero over label shuffled data \cite{zhang2016understanding} can be seen as a result of exploiting $\mathbf{x}_h$ and overfitting sample-specific idiosyncrasy. 
We will discuss more in the following sections. 

\subsection{Trade-off between Robustness and Accuracy}
We continue with Remark~\ref{remark:main} and discuss CNN's trade-off 
between robustness and accuracy given $\theta$
from the image frequency perspective. 
We first formally state the accuracy of $\theta$ as:
\begin{align}
    \mathbb{E}_{(\mathbf{x}, \mathbf{y})}\alpha (f(\mathbf{x};\theta), \mathbf{y})
\label{eq:accu}
\end{align}
and the adversarial robustness of $\theta$ as in \textit{e.g.}, \cite{carlini2019evaluating}: 
\begin{align}
    \mathbb{E}_{(\mathbf{x}, \mathbf{y})} \min_{\mathbf{x}':d(\mathbf{x}',\mathbf{x}) \leq \epsilon} \alpha (f(\mathbf{x}';\theta), \mathbf{y})
\label{eq:robust}
\end{align}
where $\epsilon$ is the upper bound of the perturbation allowed. 

With another assumption (referred to as A2): ``for model $\theta$, there exists a sample $\langle \mathbf{x}, \mathbf{y} \rangle$ such that: 
\begin{align*}
    f(\mathbf{x};\theta) \neq f(\mathbf{x}_l;\theta), \textnormal{''}
\end{align*}
we can extend our main argument (Remark~\ref{remark:main}) to a formal statement: 
\begin{corollary}
\label{theorem:main}
With assumptions A1 and A2, there exists a sample $\langle \mathbf{x}, \mathbf{y} \rangle$ that the model $\theta$ cannot predict both accurately (evaluated to be 1.0 by Equation~\ref{eq:accu}) and robustly (evaluated to be 1.0 by Equation~\ref{eq:robust}) under any distance metric $d(\cdot,\cdot)$ and bound $\epsilon$ as long as $\epsilon \geq d(\mathbf{x},\mathbf{x}_l)$.
\end{corollary}


The proof is a direct outcome of the previous discussion and thus omitted. 
The Assumption A2 can also be verified empirically (\textit{e.g.}, in Figure~\ref{fig:intro}), 
therefore we can safely state that Corollary~\ref{theorem:main} can serve as one of the explanations to the trade-off between CNN's robustness and accuracy. 


\section{Rethinking Data before Rethinking Generalization}
\label{sec:rethinking}
\subsection{Hypothesis}
Our first aim is to offer some intuitive explanations 
to the empirical results observed in \cite{zhang2016understanding}: 
neural networks can easily fit label-shuffled data. 
While we have no doubts that neural networks are capable of 
memorizing the data due to its capacity, 
the interesting question arises:
``if a neural network can easily memorize the data, why it cares to learn the generalizable patterns out of the data, in contrast to directly memorizing everything to reduce the training loss?''

Within the perspective introduced in Remark 1, 
our hypothesis is as follows:
Despite the same outcome as a minimization of the training loss, 
the model considers different level of features in the two situations:
\begin{itemize}
    \item In the original label case, the model will first pick up \lfc{}, then gradually pick up the \hfc{} to achieve higher training accuracy. 
    \item In the shuffled label case, as the association between \lfc{} and the label is erased due to shuffling, the model has to memorize the images when the \lfc{} and \hfc{} are treated equally. 
\end{itemize}


\subsection{Experiments}
\label{sec:rethinking:exp}
\begin{figure}
    \centering
    \includegraphics[width=0.40\textwidth]{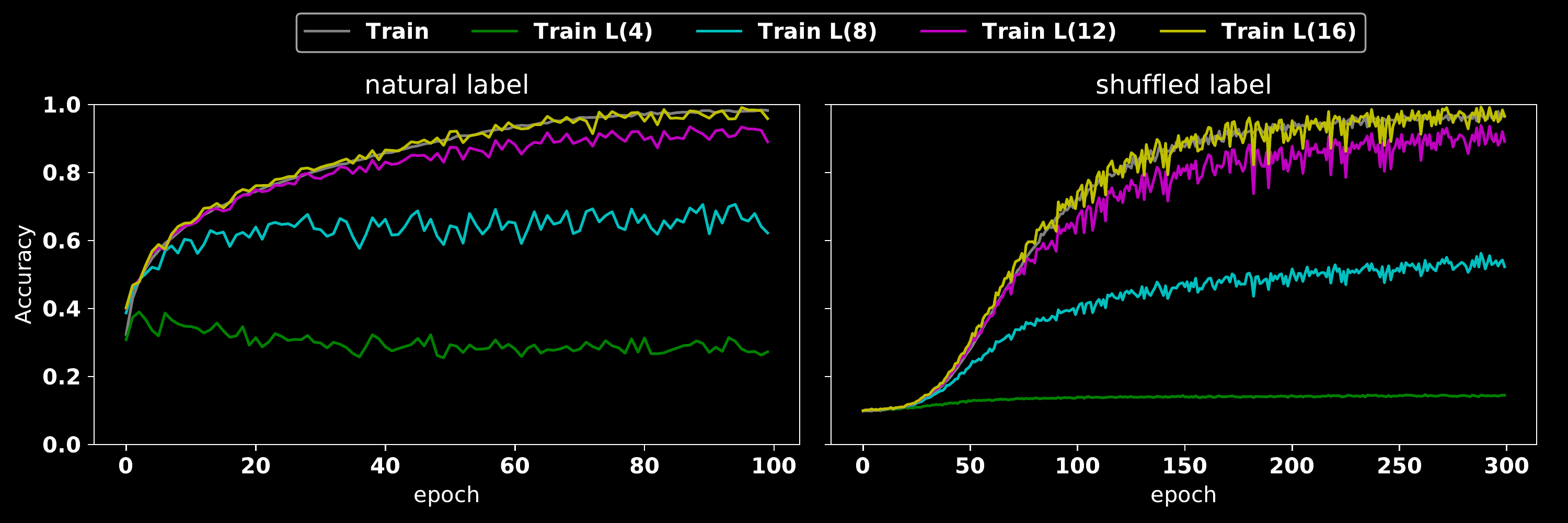}
    \caption{Training curves of the original label case (100 epoches) and shuffled label case (300 epoches), together plotted with the low-frequent counterpart of the images.
    All curves in this figure are from train samples. }
    \label{fig:rethinking}
\end{figure}

We set up the experiment to test our hypothesis. 
We use ResNet-18 \cite{He_2016_CVPR} for CIFAR10 dataset \cite{krizhevsky2009learning} as the base experiment. 
The vanilla set-up, which we will use for the rest of this paper, 
is to run the experiment with 100 epoches with the ADAM optimizer \cite{kingma2014adam} with learning rate set to be $10^{-4}$ and batch size set to be $100$, when weights are initialized with Xavier initialization \cite{pmlr-v9-glorot10a}. 
Pixels are all normalized to be $[0, 1]$. 
All these experiments are repeated in MNIST \cite{deng2012mnist}, FashionMNIST \cite{xiao2017fashion}, and a subset of ImageNet \cite{imagenet_cvpr09}. These efforts are reported in the Appendix. 
We train two models, with the natural label setup and the shuffled label setup, denote as \mn{} and \ms{}, respectively; 
the \ms{} needs 300 epoches to reach a comparative training accuracy. 
To test which part of the information the model picks up, 
for any $\mathbf{x}$ in the training set, 
we generate the low-frequency counterparts $\mathbf{x}_l$ with $r$ set to $4$, $8$, $12$, $16$ respectively.
We test the how the training accuracy changes for these low-frequency data collections along the training process. 

The results are plotted in Figure~\ref{fig:rethinking}. 
The first message is the \ms{} takes a longer training time than \mn{} to reach the same training accuracy (300 epoches vs. 100 epoches), 
which suggests that memorizing the samples as an ``unnatural'' behavior in contrast to learning the generalizable patterns. 
By comparing the curves of the low-frequent training samples, we notice that \mn{} learns more of the low-frequent patterns (\textit{i.e.}, when $r$ is $4$ or $8$) than \ms{}. 
Also, \ms{} barely learns any \lfc{} when $r=4$, 
while on the other hand, even at the first epoch, 
\mn{} already learns around 40\% of the correct 
\lfc{} when $r=4$. 
This disparity suggests that when \mn{} prefers to pick up the \lfc{}, 
\ms{} does not have a preference between \lfc{} vs. \hfc{}. 

If a model can exploit multiple different sets of signals, 
then why \mn{} prefers to learn \lfc{}
that happens to align well with the human perceptual preference? 
While there are explanations suggesting 
neural networks' tendency 
towards simpler functions \cite{rahaman2018spectral}, 
we conjecture that this is simply because, 
since the data sets are organized and annotated by human,
the \lfc{}-label association is more ``generalizable''
than the one of \hfc{}:
picking up \lfc{}-label association will lead to the steepest descent of the loss surface, especially at the early stage of the training. 

\begin{table}[]
\centering
\footnotesize
\caption{We test the generalization power of \lfc{} and \hfc{} by training the model with $\mathbf{x}_l$ or $\mathbf{x}_h$ and test on the original test set. }
\begin{tabular}{ccc|ccc}
\hline
\multicolumn{3}{c}{\lfc{}} & \multicolumn{3}{|c}{\hfc{}} \\ \hline
$r$ & train acc. & test acc. & $r$ & train acc. & test acc.\\
4 & 0.9668 & 0.6167 & 4 & 0.9885 & 0.2002 \\
8 & 0.9786 & 0.7154 & 8 & 0.9768 & 0.092 \\
12 & 0.9786 & 0.7516 & 12 & 0.9797 & 0.0997 \\
16 & 0.9839 & 0.7714 & 16 & 0.9384 & 0.1281 \\ \hline
\end{tabular}
\label{tab:rethinking}
\end{table}

To test this conjecture, we repeat the experiment of \mn{}, 
but instead of the original train set, 
we use the $\mathbf{x}_l$ or $\mathbf{x}_h$ 
(normalized to have the standard pixel scale)
and test how well the model can perform on 
original test set. 
Table~\ref{tab:rethinking} suggests that
\lfc{} is much more ``generalizable''
than \hfc{}. 
Thus, it is not surprising 
if a model first picks up 
\lfc{} as it leads to the steepest descent of the loss surface.

\subsection{A Remaining Question}
Finally, 
we want to raise a question: 
The coincidental alignment between networks' preference in \lfc{} 
and human perceptual preference 
might be a simple result of the ``survival bias'' 
of the many technologies invented 
one of the other
along the process of climbing the ladder of 
the state-of-the-art. 
In other words, 
the almost-100-year development process of neural networks functions like 
a ``natural selection'' of technologies \cite{wang2017origin}. 
The survived ideas 
may happen to match the human preferences, 
otherwise, the ideas may not even be published 
due to the incompetence in climbing the ladder. 

However, an interesting question will be how well these
ladder climbing techniques align with the human visual preference. 
We offer to evaluate these techniques
with our frequency tools.


\section{Training Heuristics}
\label{sec:heuristics}
We continue to reevaluate
the heuristics that helped in climbing the ladder of state-of-the-art accuracy. 
We evaluate these heuristics to test
the generalization performances towards
\lfc{} and \hfc{}. 
Many renowned techniques
in the ladder of accuracy
seem to exploit \hfc{} more or less. 

\subsection{Comparison of Different Heuristics} 
We test multiple heuristics by inspecting 
the prediction accuracy over \lfc{} and \hfc{} 
with multiple choices of $r$
along the training process
and plot the training curves. 





\begin{figure*}
    \centering
    \includegraphics[width=0.9\textwidth]{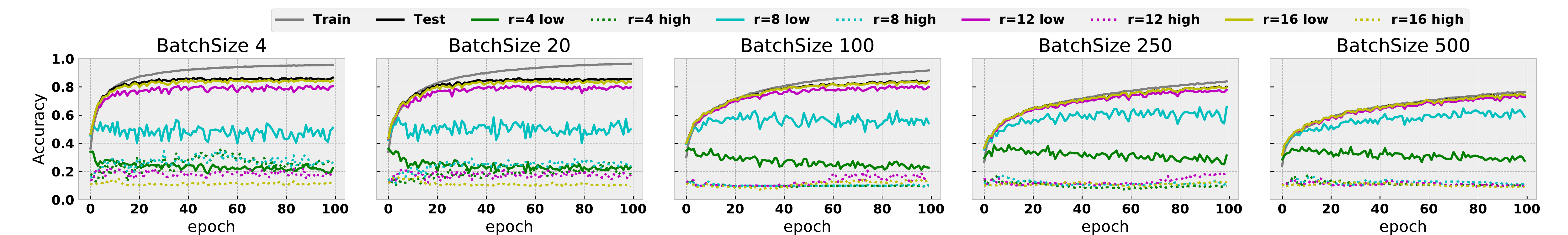}
    \caption{Plots of accuracy of different epoch sizes along the epoches for train, test data, as well as \lfc{} and \hfc{} with different radii.}
    \label{fig:epoch}
\end{figure*}

\textbf{Batch Size:}
We then investigate how the choices of batch size 
affect the generalization behaviors. 
We plot the results in Figure~\ref{fig:epoch}. 
As the figure shows, 
smaller batch size appears to excel in improving training and testing accuracy, 
while bigger batch size seems to stand out in closing the generalization gap. 
Also, it seems the generalization gap is closely related 
to the model's tendency in capturing \hfc{}:
models trained with bigger epoch sizes are more invariant to \hfc{} 
and introduce smaller differences in training accuracy and testing accuracy.
The observed relation is intuitive because the smallest generalization gap will be achieved once the model behaves like a human (because it is the human who annotate the data). 

The observation in Figure~\ref{fig:epoch} also chips in the discussion in the previous section about ``generalizable'' features. 
Intuitively, with bigger epoch size, the features that can 
lead to steepest descent of the loss surface  
are more likely to be the ``generalizable'' patterns of the data, 
which are \lfc{}. 

\begin{figure*}
    \centering
    \includegraphics[width=0.9\textwidth]{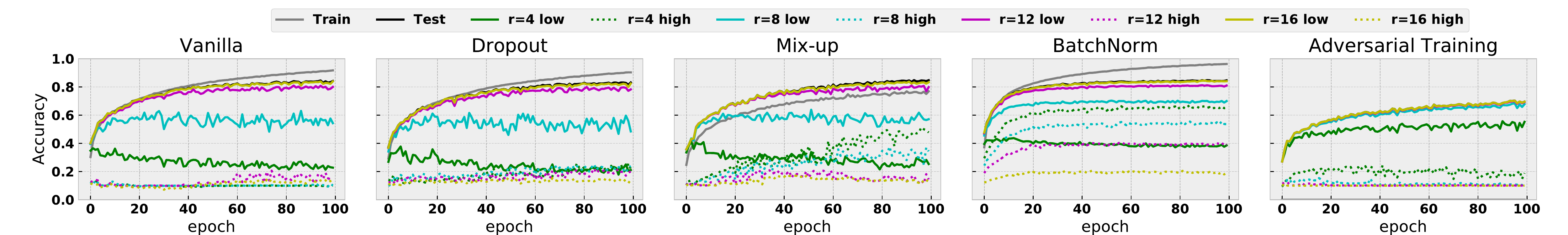}
    \caption{Plots of accuracy of different heuristics along the epoches for train, test data, as well as \lfc{} and \hfc{} with different radii.}
    \label{fig:heuristics}
\end{figure*}

\textbf{Heuristics:}
We also test how different training methods react to \lfc{} and \hfc{}, including
\begin{itemize}
    \item Dropout \cite{hinton2016system}: A heuristic that drops weights randomly during training. 
    We apply dropout on fully-connected layers with $p=0.5$. 
    \item Mix-up \cite{zhang2017mixup}: A heuristic that linearly integrate samples and their labels during training. 
    We apply it with standard hyperparameter $\alpha=0.5$. 
    \item BatchNorm \cite{ioffe2015batch}: A method that perform the normalization for each training mini-batch to accelerate Deep Network training process. It allows us to use a much higher learning rate and reduce overfitting, similar with Dropout. 
    We apply it with setting scale $\gamma$ to 1 and offset $\beta$ to 0.
    \item Adversarial Training \cite{madry2017towards}: A method that augments the data through adversarial examples generated by a threat model during training. 
    It is widely considered as one of the most successful adversarial robustness (defense) method. 
    Following the popular choice, we use PGD with $\epsilon=8/255$ ($\epsilon=0.03$ ) as the threat model. 
\end{itemize}

We illustrate the results in Figure~\ref{fig:heuristics}, where the first panel is the vanilla set-up, and then each one of the four heuristics are tested in the following four panels. 


Dropout roughly behaves similarly to the vanilla set-up in our experiments. 
Mix-up delivers a similar prediction accuracy, however, 
it catches much more \hfc{}, which is probably not surprising 
because the mix-up augmentation does not encourage anything about \lfc{} explicitly, 
and the performance gain is likely due to attention towards \hfc{}. 

Adversarial training mostly behaves as expected:
it reports a lower prediction accuracy, 
which is likely due to the trade-off between robustness and accuracy.
It also reports a smaller generalization gap, 
which is likely as a result of picking up 
``generalizable'' patterns, 
as verified by its invariance towards \hfc{} (\textit{e.g.}, $r=12$ or $r=16$). 
However, adversarial training 
seems to be sensitive to the \hfc{} when $r=4$,
which is ignored even by the vanilla set-up. 

The performance of BatchNorm is notable:
compared to the vanilla set-up, BatchNorm 
picks more information in both \lfc{} and \hfc{}, 
especially when $r=4$ and $r=8$.
This BatchNorm's tendency in capturing 
\hfc{} is also related to observations 
that BatchNorm encourages adversarial 
vulnerability \cite{galloway2019batch}. 

\textbf{Other Tests:}
We have also tested other heuristics or methods by only changing along one dimension while the rest is fixed the same as the vanilla set-up in Section~\ref{sec:rethinking}. 
    
Model architecture: We tested LeNet \cite{lecun1998gradient}, AlexNet \cite{krizhevsky2012imagenet}, VGG \cite{simonyan2014very}, and ResNet \cite{he2016deep}. The ResNet architecture seems advantageous toward previous inventions
at different levels:
it reports better vanilla test accuracy, 
smaller generalization gap (difference between training and testing accuracy),
and a weaker tendency in capturing \hfc{}. 

Optimizer: We tested SGD, ADAM \cite{kingma2014adam}, AdaGrad \cite{duchi2011adaptive}, AdaDelta \cite{zeiler2012adadelta}, and RMSprop. 
We notice that SGD seems to be the only one suffering from the tendency towards significantly capturing \hfc{}, while the rest are on par within our experiments. 

\subsection{A hypothesis on Batch Normalization}
Based on the observation, we hypothesized that 
one of BatchNorm's advantage is,
through normalization, 
to align the distributional disparities
of different predictive signals. 
For example, 
\hfc{} usually shows smaller magnitude
than \lfc{}, 
so a model trained without BatchNorm may not easily 
pick up these \hfc{}. 
Therefore, the higher convergence speed
may also be considered as a direct 
result of capturing different predictive
signals simultaneously. 

\begin{figure}
    \centering
    \includegraphics[width=0.4\textwidth]{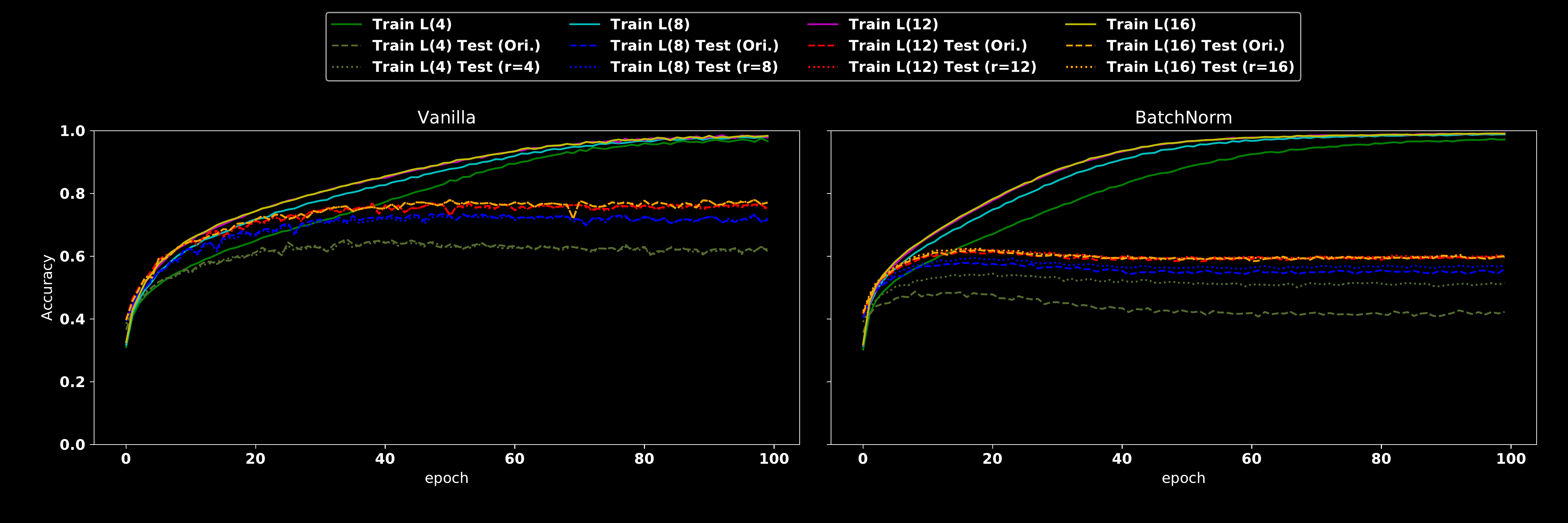}
    \caption{Comparison of models with vs. without BatchNorm trained with \lfc{} data.}
    \label{fig:batchNorm}
\end{figure}

To verify this hypothesis, 
we compare the performance of models trained with vs. without BatchNorm 
over \lfc{} data and plot the results in Figure~\ref{fig:batchNorm}. 

As Figure~\ref{fig:batchNorm} shows, when the model is trained with only \lfc{}, 
BatchNorm does not always help improve the predictive performance, 
either tested by original data or by corresponding \lfc{} data.
Also, the smaller the radius is, 
the less the BatchNorm helps. 
Also, in our setting, 
BatchNorm does not generalize as well as the vanilla setting,
which may raise a question about the benefit of BatchNorm. 

However, BatchNorm still seems to at least 
boost the convergence of training accuracy. 
Interestingly, 
the acceleration is the smallest when $r=4$. 
This observation further aligns with our hypothesis: 
if one of BatchNorm's advantage is to encourage the model to capture different
predictive signals, 
the performance gain of BatchNorm is the most limited 
when the model is trained with \lfc{} when $r=4$.  


\section{Adversarial Attack \& Defense}
\label{sec:adversarial}
As one may notice, 
our observation of \hfc{} can be directly linked to the phenomenon of
``adversarial example'':
if the prediction relies on \hfc{}, 
then perturbation of \hfc{} will significantly 
alter the model's response, 
but such perturbation may not be observed to human at all,
creating the unintuitive behavior of neural networks. 


This section is devoted to study the relationship between 
adversarial robustness and model's tendency in exploiting \hfc{}. 
We first discuss the linkage between
the ``smoothness'' of convolutional kernels 
and model's sensitivity towards \hfc{} (\cref{sec:adv:main}), 
which serves the tool for our follow-up analysis. 
With such tool, 
we first show that
adversarially robust models tend to have ``smooth'' kernels (\cref{sec:adv:pgd}), 
and then demonstrate that 
directly smoothing the kernels (without training) can help improve the 
adversarial robustness towards some attacks (\cref{sec:adv:smooth}). 


\subsection{Kernel Smoothness vs. Image Frequency}
\label{sec:adv:main}


 


As convolutional theorem \cite{bracewell1986fourier} states,
the convolution operation of images is equivalent to the element-wise multiplication of image frequency domain. 
Therefore, roughly, if a convolutional kernel has negligible weight at the high-end of the frequency domain, it will weigh \hfc{} accordingly.
This may only apply to the convolutional kernel at the first layer 
because the kernels at higher layer do not directly with the data,
thus the relationship is not clear. 

Therefore, we argue that, to push the model to ignore the \hfc{}, 
one can consider to force the model to learn the convolutional kernels
that have only negligible weights at the high-end of the frequency domain. 

Intuitively (from signal processing knowledge), if the convolutional kernel is ``smooth'', 
which means that there is no dramatics fluctuations 
between adjacent weights, 
the corresponding frequency domain will see a 
negligible amount of high-frequency signals. 
The connections have been mathematically proved \cite{platonov2005fourier,titchmarsh1948introduction}, 
but these proved exact relationships are out of the scope of this paper. 

\subsection{Robust Models Have Smooth Kernels}
\label{sec:adv:pgd}

\begin{figure}
    \centering
    \subfloat[convoluational kernels of \mn{} ]{\includegraphics[width=0.4\textwidth]{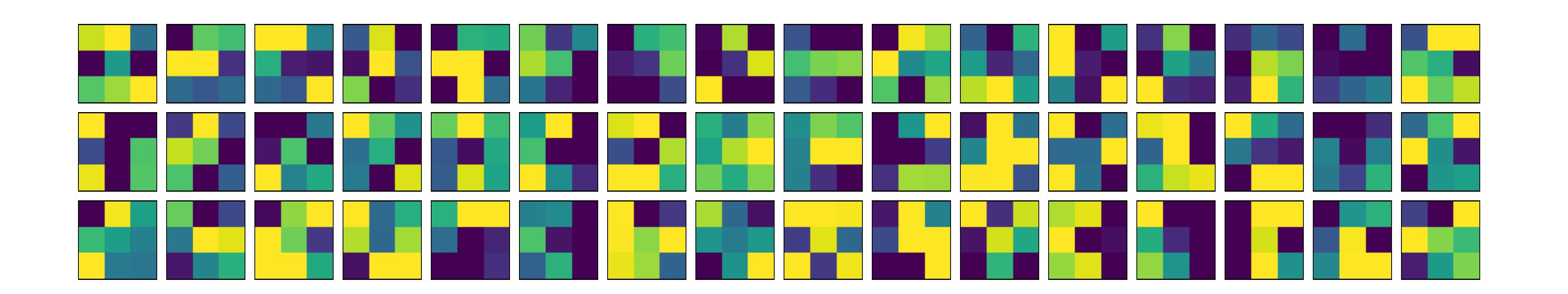}}\\
    \subfloat[convoluational kernels of \ma{} ]{\includegraphics[width=0.4\textwidth]{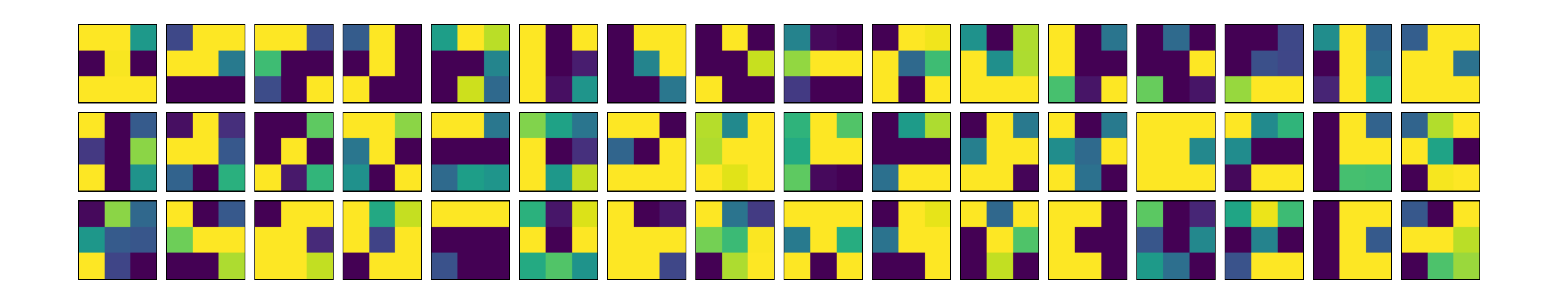}}\\
    \subfloat[convoluational kernels of \mn{}($\rho$=1.0) ]{\includegraphics[width=0.4\textwidth]{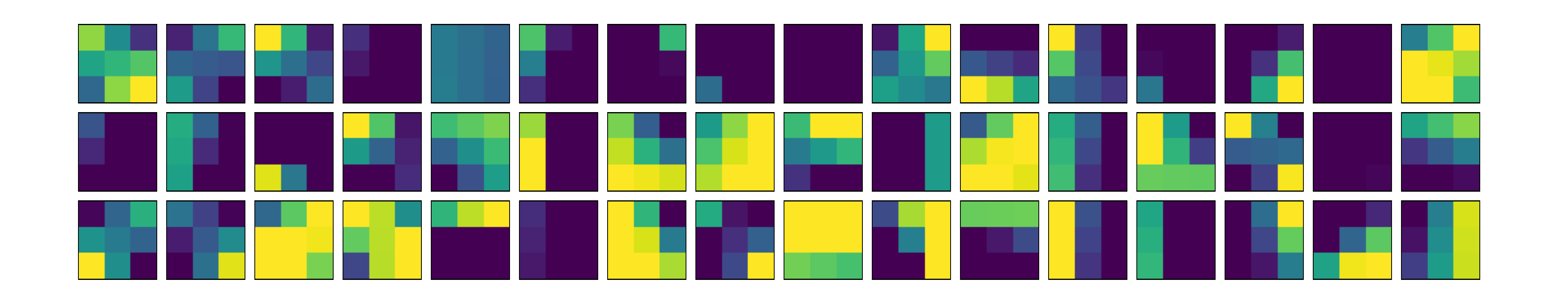}}\\
    \subfloat[convoluational kernels of \ma{}($\rho$=1.0) ]{\includegraphics[width=0.4\textwidth]{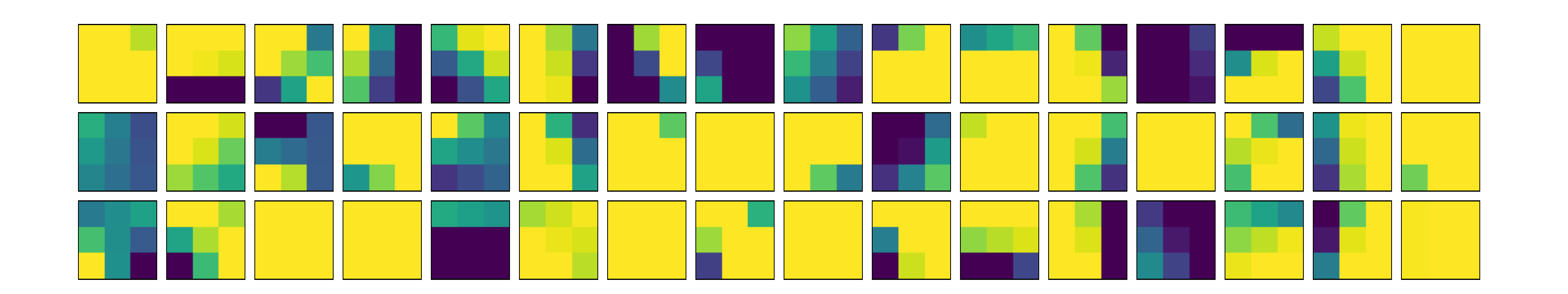}}\\
    \caption{Visualization of convolutional kernels (16 kernels each channel $\times$ 3 channels at the first layer) of models.}
    \label{fig:kernel}
\end{figure}

To understand the connection between ``smoothness'' and adversarial robustness, 
we visualize the convolutional kernels at the first layer
of the models trained in the vanilla manner (\mn{})
and trained with adversarial training (\ma{}) in Figure~\ref{fig:kernel} (a) and (b).

Comparing Figure~\ref{fig:kernel}(a) and Figure~\ref{fig:kernel}(b), 
we can see that the kernels of \ma{} tend to show a more smooth pattern, 
which can be observed by noticing that
the adjacent weights of kernels of \ma{} tend to share the same color. 
The visualization may not be very clear because the convolutional kernel is only [3 $\times$ 3] in ResNet, 
the message is delivered more clearly in Appendix with other architecture when the first layer has kernel of the size [5 $\times$ 5]. 

\begin{table*}[t!]
\centering 
\footnotesize
\begin{tabular}{cccccccc}
\hline
\multirow{2}{*}{} & \multirow{2}{*}{Clean} & \multicolumn{3}{c}{FGSM} & \multicolumn{3}{c}{PGD} \\
 &  & $\epsilon=0.03$ & $\epsilon=0.06$ & $\epsilon=0.09$ & $\epsilon=0.03$ & $\epsilon=0.06$ & $\epsilon=0.09$ \\ \hline
\mn{} & \textbf{0.856} & 0.107 & 0.069 & 0.044 & 0.003 & 0.002 & 0.002 \\
\mn{}($\rho=0.10$) & 0.815 & 0.149 & 0.105 & 0.073 & 0.009 & 0.002 & 0.001 \\
\mn{}($\rho=0.25$) & 0.743 & 0.16 & 0.11 & 0.079 & 0.021 & 0.005 & 0.005 \\
\mn{}($\rho=0.50$) & 0.674 & 0.17 & 0.11 & 0.083 & 0.031 & 0.016 & 0.014 \\
\mn{}($\rho=1.0$) & 0.631 & \textbf{0.171} & \textbf{0.14} & \textbf{0.127} & \textbf{0.086} & \textbf{0.078} & \textbf{0.078} \\ \hline
\ma{} & \textbf{0.707} & \textbf{0.435} & \textbf{0.232} & 0.137 & \textbf{0.403} & 0.138 & 0.038 \\
\ma{}($\rho=0.10$) & 0.691 & 0.412 & 0.192 & 0.109 & 0.379 & 0.13 & 0.047 \\
\ma{}($\rho=0.25$) & 0.667 & 0.385 & 0.176 & 0.097 & 0.352 & 0.116 & 0.04 \\
\ma{}($\rho=0.50$) & 0.653 & 0.365 & 0.18 & 0.106 & 0.334 & 0.121 & 0.062 \\
\ma{}($\rho=1.0$) & 0.638 & 0.356 & 0.223 & \textbf{0.186} & 0.337 & \textbf{0.175} & \textbf{0.131} \\ \hline
\end{tabular}
\caption{Prediction performance of models against different adversarial attacks with different $\epsilon$.}
\label{tab:adversarial}
\end{table*}

\subsection{Smoothing Kernels Improves Adversarial Robustness}
\label{sec:adv:smooth}

The intuitive argument in \cref{sec:adv:main} 
and empirical findings in \cref{sec:adv:pgd} directly lead to a question
of whether we can improve the adversarial robustness 
of models by smoothing the convolutional kernels 
at the first layer. 

Following the discussion, 
we introduce an extremely simple method that appears to 
improve the adversarial robustness against FGSM \cite{goodfellow2015explaining} and PGD \cite{kurakin2016adversarial}.
For a convolutional kernel $\mathbf{w}$, 
we use $i$ and $j$ to denote its column and row indices, 
thus $\mathbf{w}_{i,j}$ denotes the value at $i$\textsuperscript{th} row 
and $j$\textsuperscript{th} column. 
If we use $\mathcal{N}(i,j)$ to denote the set of the spatial neighbors of $(i,j)$, 
our method is simply:
\begin{align}
    \mathbf{w}_{i,j} = \mathbf{w}_{i,j} + \sum_{(h,k) \in \mathcal{N}(i,j)}\rho \mathbf{w}_{h,k},
    \label{eq:method}
\end{align}
where $\rho$ is a hyperparameter of our method. 
We fix $\mathcal{N}(i,j)$ to have eight neighbors. 
If $(i,j)$ is at the edge, then we simply generate the out-of-boundary values by duplicating the values on the boundary. 

In other words, 
we try to smooth the kernel through simply 
reducing the adjacent differences by mixing the adjacent values. 
The method barely has any computational load, 
but appears to improve the adversarial robustness 
of \mn{} and \ma{} 
towards FGSM and PGD, 
even when \ma{} is trained with PGD as the threat model. 

In Figure~\ref{fig:kernel}, 
we visualize the convolutional kernels with 
our method applied to \mn{} and \ma{} with
$\rho=1.0$, 
denoted as \mn{}($\rho=1.0$) and \ma{}($\rho=1.0$), respectively.
As the visualization shows, 
the resulting kernels tend to show a significantly smoother pattern. 

We test the robustness of the models smoothed by our method against 
FGSM and PGD with different choices of $\epsilon$, where the maximum of perturbation is 1.0. 
As Table~\ref{tab:adversarial} shows, 
when our smoothing method is applied, 
the performance of clean accuracy directly plunges, 
but the performance of adversarial robustness improves. 
In particular, our method helps 
when the perturbation is allowed to be relatively large. 
For example, when $\epsilon=0.09$ (roughly $23/255$), 
\mn{}($\rho=1.0$) even outperforms \ma{}. 
In general, our method can easily improve the 
adversarial robustness of \mn{}, 
but can only improve upon \ma{} 
in the case where $\epsilon$ is larger, 
which is probably because the \ma{} is trained 
with PGD($\epsilon=0.03$) as the threat model.



\section{Beyond Image Classification}
\label{sec:beyond}
We aim to explore more than image classification tasks. 
We investigate 
in the object detection task. 
We use RetinaNet \cite{lin2017focal} with ResNet50 \cite{he2016deep} + FPN \cite{lin2017feature} as the backbone. 
We train the model with COCO detection train set \cite{lin2014microsoft}
and perform inference in its validation set, which includes 5000 images, and achieve an MAP of $35.6\%$. 

Then we choose $r=128$ and maps the images into $\mathbf{x}_l$ and $\mathbf{x}_h$ and test with the same model and get $27.5\%$ MAP with \lfc{} and $10.7\%$ MAP with \hfc{}.
The performance drop from $35.6\%$ to $27.5\%$ intrigues us 
so we further study whether the same drop 
should be expected from human. 

\subsection{Performance Drop on LFC}

The performance drop from the $\mathbf{x}$ to $\mathbf{x}_l$ may be expected 
because $\mathbf{x}_l$ may not have the rich information from the original images 
when \hfc{} are dropped. 
In particular, 
different from image classification, 
\hfc{} may play a significant role in depicting some objects, 
especially the smaller ones. 


\begin{figure}
    \centering
    \includegraphics[width=.225\linewidth]{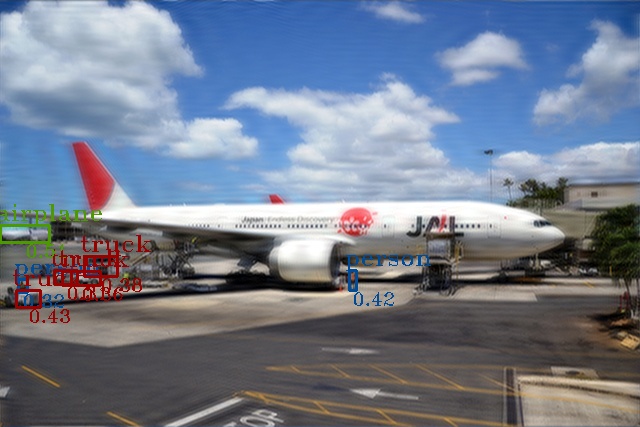}
    \,
    \includegraphics[width=.225\linewidth]{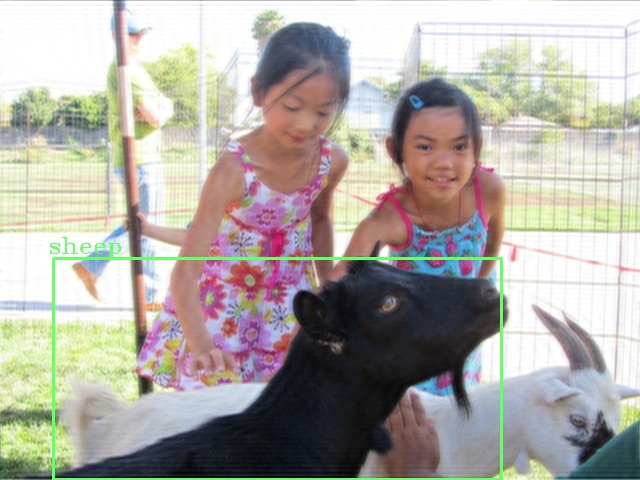}
    \,
    \includegraphics[width=.225\linewidth]{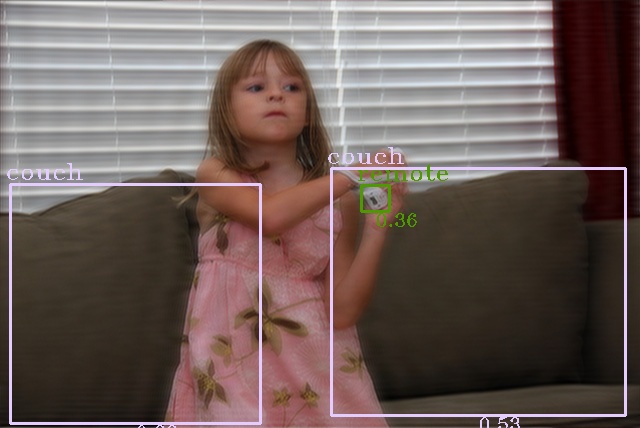}
    \,
    \includegraphics[width=.225\linewidth]{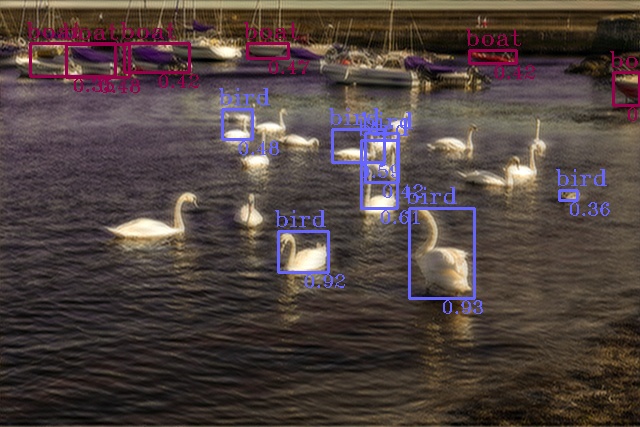}
    \caption{Some objects are recognized worse (lower MAP scores) when the experiments are repeated with low-frequent images. Marked objects are the ones that induce differences.}
    \label{fig:object:drop}
\end{figure}

Figure~\ref{fig:object:drop} illustrates a few examples, where some objects are recognized worse in terms of MAP scores when the input images are replaced by the low-frequent counterparts. 
This disparity may be expected because the low-frequent images tend to be blurry 
and some objects may not be clear to a human either 
(as the left image represents). 

\subsection{Performance Gain on LFC}

However, the disparity gets interesting
when we inspect the performance gap in the opposite direction. 
We identified 1684 images that
for each of these images, 
the some objects are recognized better (high MAP scores) 
in comparison to the original images. 


\begin{figure}
    \centering
    \includegraphics[width=.225\linewidth]{objectionDetectionComparison/sfig1.jpg}
    \,
    \includegraphics[width=.225\linewidth]{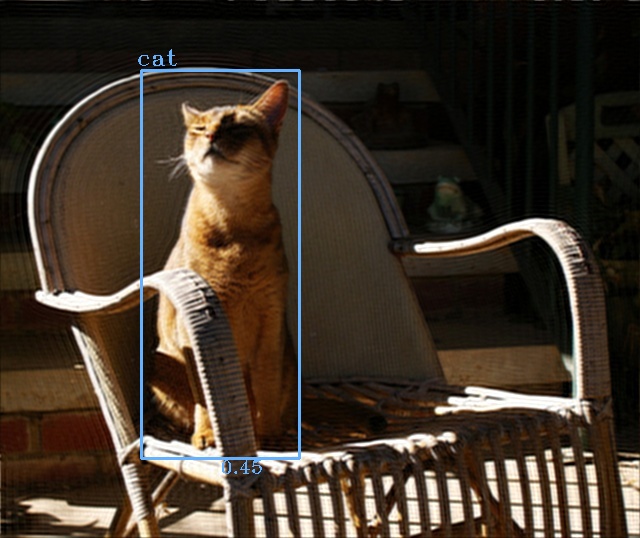}
    \,
    \includegraphics[width=.225\linewidth]{objectionDetectionComparison/sfig3.jpg}
    \,
    \includegraphics[width=.225\linewidth]{objectionDetectionComparison/sfig2.jpg}
    \caption{Some objects are recognized better (higher MAP scores) when the experiments are repeated with low-frequent images. Marked objects are the ones that induce differences.}
    \label{fig:object:gain}
\end{figure}

The results are shown in Figure~\ref{fig:object:gain}. 
There seems 
no apparent reasons why 
these objects are recognized better in low-frequent images, 
when inspected by human. 
These observations strengthen our argument in the perceptual disparity 
between CNN and human also exist in more advanced computer vision tasks other than image classification. 

\section{Discussion: Are HFC just Noises?}
\label{sec:discussion}
To answer this question\blfootnote{Code used in the paper: \href{https://github.com/HaohanWang/HFC}{https://github.com/HaohanWang/HFC}}, we experiment with another frequently used image denoising method: truncated singular value decomposition (SVD). 
We decompose the image and separate the image into one reconstructed with dominant singular values and one with trailing singular values. 
With this set-up, we find much fewer images supporting the story in Figure~\ref{fig:intro}.
Our observations suggest the signal CNN exploit is more than just random ``noises''. 






\section{Conclusion \& Outlook}
\label{sec:con}
We investigated how image frequency spectrum 
affects the generalization behavior of CNN, 
leading to multiple interesting explanations 
of the generalization behaviors of neural networks
from a new perspective: 
there are multiple signals in the data, 
and not all of them align with human's visual preference. 
As the paper comprehensively covers many topics, 
we briefly reiterate the main lessons learned: 
\begin{itemize}
    \item CNN may capture \hfc{} that are misaligned with human visual preference (\cref{sec:method}), resulting in generalization mysteries such as the paradox of learning label-shuffled data (\cref{sec:rethinking}) and adversarial vulnerability (\cref{sec:adversarial}). 
    \item Heuristics that improve accuracy (\textit{e.g.}, Mix-up and BatchNorm) may encourage capturing \hfc{} (\cref{sec:heuristics}). Due to the trade-off between accuracy and robustness (\cref{sec:method}), we may have to rethink the value of them. 
    \item Adversarially robust models tend to have smooth convolutional kernels, the reverse is not always true (\cref{sec:adversarial}).  
    \item Similar phenomena are noticed in the context of object detection (\cref{sec:beyond}), with more conclusions yet to be drawn.  
\end{itemize}

Looking forward, we hope our work serves as a call towards future era of computer vision research, 
where the state-of-the-art is not as important as we thought. 
\begin{itemize}
    \item A single numeric on the leaderboard, 
    while can significantly boost the research towards a direction, 
    does not reliably reflect the alignment between models and human, 
    while such an alignment is arguably paramount. 
    \item We hope our work will set forth towards 
    a new testing scenario where the performance of low-frequent counterparts 
    needs to be reported together with the performance of the original images. 
    \item Explicit inductive bias considering how a human views the data (\textit{e.g.}, \cite{wang2018learning,wang2019learning}) may play a significant role in the future. 
    In particular, neuroscience literature have shown that 
    human tend to rely on low-frequent signals in recognizing objects \cite{awasthi2011faster,bar2004visual}, which may inspire development of future methods. 
\end{itemize}

{\small
\section*{Acknowledgements}
\setlength\parindent{0pt}
This material is based upon work supported by NIH R01GM114311, NIH P30DA035778, and NSF IIS1617583. Any opinions, findings and conclusions or recommendations expressed in this material are those of the author(s) and do not necessarily reflect the views of the National Institutes of Health or the National Science Foundation.
}

{\small
\bibliographystyle{ieee_fullname}
\bibliography{ref}
}

\end{document}